\documentclass[jair,twoside,11pt,theapa]{article}
\usepackage[table,xcdraw]{xcolor}
\definecolor{royalblue}{rgb}{0.25, 0.41, 0.88}

\usepackage[colorlinks=true,linkcolor=black,citecolor=black,urlcolor=royalblue]{hyperref}
\usepackage{url}
\usepackage{jair, theapa, rawfonts}
\ShortHeadings{MORL/D: A Taxonomy and Framework}
{Felten, Talbi \& Danoy}
\firstpageno{1}

\jairheading{X}{2024}{1-44}{03/2023}{??/2024}
\usepackage[pdftex]{graphicx}

\usepackage{amsmath}
\usepackage{amssymb}
\usepackage{algorithm}
\usepackage{algpseudocode}
\usepackage{subcaption}
\usepackage{lipsum}
\usepackage{lscape}
\usepackage{afterpage}

\newcommand{\Ex}{\mathbb{E}}

\usepackage{graphicx}

\usepackage{bbm}
\usepackage{caption}

\usepackage{subcaption}

\usepackage{multirow}
\usepackage{breakcites}
\DeclareMathOperator*{\argmax}{argmax}
\DeclareMathOperator*{\argg}{arg}
\definecolor{myblue}{HTML}{4249ff}

\title{Multi-Objective Reinforcement Learning Based on Decomposition: A Taxonomy and Framework}
\author{\name Florian Felten \email florian.felten@uni.lu \\
\addr SnT, University of Luxembourg
\AND
El-Ghazali Talbi \email el-ghazali.talbi@univ-lille.fr\\
\addr CNRS/CRIStAL, University of Lille \\
\addr FSTM/DCS, University of Luxembourg
\AND
Grégoire Danoy \email gregoire.danoy@uni.lu \\
\addr FSTM/DCS, University of Luxembourg\\
\addr SnT, University of Luxembourg \\
}

\begin{document}

\maketitle
\begin{abstract}

Multi-objective reinforcement learning (MORL) extends traditional RL by seeking policies making different compromises among conflicting objectives. The recent surge of interest in MORL has led to diverse studies and solving methods, often drawing from existing knowledge in multi-objective optimization based on decomposition (MOO/D). Yet, a clear categorization based on both RL and MOO/D is lacking in the existing literature. Consequently, MORL researchers face difficulties when trying to classify contributions within a broader context due to the absence of a standardized taxonomy. 
To tackle such an issue, this paper introduces multi-objective reinforcement learning based on decomposition (MORL/D), a novel methodology bridging the literature of RL and MOO. A comprehensive taxonomy for MORL/D is presented, providing a structured foundation for categorizing existing and potential MORL works. The introduced taxonomy is then used to scrutinize MORL research, enhancing clarity and conciseness through well-defined categorization. Moreover, a flexible framework derived from the taxonomy is introduced. This framework accommodates diverse instantiations using tools from both RL and MOO/D. Its versatility is demonstrated by implementing it in different configurations and assessing it on contrasting benchmark problems. Results indicate MORL/D instantiations achieve comparable performance to current state-of-the-art approaches on the studied problems. By presenting the taxonomy and framework, this paper offers a comprehensive perspective and a unified vocabulary for MORL. This not only facilitates the identification of algorithmic contributions but also lays the groundwork for novel research avenues in MORL.
\end{abstract}

\section{Introduction}
In the complex landscape of real-life decision-making, individuals often find themselves facing the challenge of balancing conflicting objectives. Consider the case of daily commuting, where one must select a mode of transportation. Each option comes with its own set of trade-offs, including factors like time, cost, and environmental impact. These choices are influenced by personal values and preferences. Multi-objective methods are designed to enhance the process of decision-making in such scenarios. When a user's preferences are known in advance (\textit{a priori}), the solution typically results in a single optimal choice. However, in cases where user preferences are uncertain or undefined, the search for the best decision leads to a set of optimal solutions, and users are asked to make informed choices afterward (\textit{a posteriori}).

Multi-objective optimization (MOO) represents a well-established field dedicated to solving such problems efficiently. Traditionally, evolutionary algorithms (EAs) have been the go-to tools for searching for solutions in MOO. Moreover, recent years have witnessed the emergence of decomposition-based methods, in which the multi-objective problem (MOP) is transformed into a set of single-objective problems (SOPs) through scalarization functions. Building on seminal work in decomposition such as MOEA/D~\shortcite{zhang_moead_2007}, a significant body of research has emerged in this area. While MOO has found successful applications in numerous problem domains, there are scenarios where the quest for solutions involves navigating a decision space that is simply too vast to explore with traditional techniques.

Notably, reinforcement learning (RL) has recently been the subject of significant research interest due to its successes in a variety of applications, which are known to be challenging for search-based methods~\shortcite{mnih_human-level_2015,silver_mastering_2016,wurman_outracing_2022}. However, this field of research is limited to agents aiming at maximizing a single objective. Thus, it has recently been expanded into more demanding scenarios, such as training the agent to learn to make compromises between multiple conflicting objectives in multi-objective RL (MORL). Given the conceptual proximity to RL and MOO, MORL researchers have explored the integration of ideas from these well-established fields to form new contributions in MORL. Although some survey papers have offered an overview of MORL's current state~\shortcite{roijers_survey_2013,hayes_practical_2022}, these surveys have primarily delved into solution concepts and theory, without comprehensively studying recent solving methods. In particular, to the best of our knowledge, no existing work has comprehensively analyzed the interactions between RL, MORL, and MOO, or systematically identified recurring patterns and approaches employed in MORL algorithms. 

Therefore, the initial portion of this work strives to clarify the similarities and distinctions between RL, MORL, and MOO, especially in scenarios where user preferences are uncertain (\textit{a posteriori}). Specifically, we aim to show that there are ways to classify and describe MORL contributions within a broader context. Hence, we present a taxonomy that aims at classifying existing and future MORL works, drawing from RL and MOO concepts. Subsequently, this taxonomy is applied to categorize existing MORL research, exemplifying its utility in comprehending the state of the art and distilling key contributions from various papers.

Building upon the theoretical foundation laid out in the taxonomy, we introduce the multi-objective reinforcement learning based on decomposition (MORL/D) framework. This modular framework can be instantiated in diverse ways using tools from both RL and MOO. To demonstrate its flexibility, we implement different variations of the framework and evaluate its performance on well-established benchmark problems. The experiments showcase how different instantiations of the framework can yield diverse results. By presenting the taxonomy and framework, our aim is to offer perspective and a shared lexicon for experts from various fields while paving the way for fresh avenues of research in MORL.

The structure of the paper unfolds as follows: Sections~\ref{sec:RL} and~\ref{sec:MOO/D} provide background information on RL and MOO, respectively. Section~\ref{sec:morld} introduces the taxonomy and the MORL/D framework, while Section~\ref{sec:using_morld} illustrates typical usages of the introduced taxonomy. In Section~\ref{sec:expe}, we demonstrate various implementations of MORL/D on benchmark problems, and the final sections encompass discussions on future directions and conclusions.

\section{Reinforcement Learning}

\label{sec:RL}
In the framework of RL, an agent interacts with an environment as follows: when faced with a particular state, the agent selects an action, which, upon execution, alters the environment, and in return, the agent receives a reward that indicates the quality of the action taken. In this context, the primary objective of the agent is to acquire a policy function, typically denoted by $\pi$, that optimally guides its action choices to maximize the cumulative rewards obtained during its interactions with the environment. After the training period, this learned policy dictates the agent's decision-making process when it encounters different states.

Formally, an RL problem is modeled by a Markov decision process (MDP), which is defined by a tuple $(\mathbb S, \mathbb A, r, p, \mu_0)$ where $\mathbb S$ are the states the agent can perceive, $\mathbb A$ are the actions the agent can undertake to interact with the environment, $r: \mathbb S \times \mathbb A \times \mathbb S \rightarrow \mathbb R$ is the reward function, $p: \mathbb S \times \mathbb A \times \mathbb S \rightarrow [0, 1]$ is the probability of transition function, giving the probability of the next state given the current state and action, and $\mu_0$ is the distribution over initial states $s_0$ \cite{sutton_reinforcement_2018}.

Within this framework, a policy $\pi: \mathbb S \times \mathbb A \rightarrow [0, 1]$ can be assigned to a numerical value after evaluation in the MDP. This evaluation value is formally referred to as the value function and can be expressed as the discounted sum of rewards collected over an infinite time horizon (or until the end of an episode) starting from a given state $s$:
\begin{equation}
v^{\pi}(s)  \equiv \Ex_{a_t \sim \pi(s_t)} \left[\sum_{t=0}^\infty \gamma^t r(s_t, a_t, s_{t+1}) | \ s_t = s \right], 
\end{equation}

\noindent where $t$ represents the timesteps at which the agent makes a choice, and $0 \leq \gamma < 1$ is the discount factor, specifying the relative importance of long-term rewards with respect to short-term rewards. Furthermore, the overall value of a policy $\pi$, regardless of the initial state, can be expressed as $v^\pi \equiv \Ex_{s_0 \sim \mu_0} v^\pi(s_0)$.
Such a value is traditionally used to define an ordering on policies, \textit{i.e.} $\pi \succeq \pi' \iff v^\pi \geq v^{\pi'}$. The goal of an RL algorithm is then to find an optimal policy $\pi^*$, defined as one which maximizes $v^\pi$: $\pi^* = \argmax_\pi v^\pi$. 

To find such an optimal policy, many RL algorithms have been published over the last decades. A high-level skeleton of the RL process is presented in Algorithm~\ref{algo:batched_td}. The algorithm first initializes its policy and an experience buffer (lines 1--2). Then, it samples experience tuples $(s_t,a_t,r_t,s_{t+1})$ from the environment by using the current policy and stores those in the buffer (lines 4--5). Optionally, to report the improvement of the policy over the training process, its value can be estimated at each iteration by computing the average returns over a predefined budget (line 6). From the buffered experiences and its current value, the policy is improved (line 7). The optimization process stops when a criterion specified by the user is met and the current policy is returned. 

\begin{figure}
    \centering
    \includegraphics[width=0.9\textwidth]{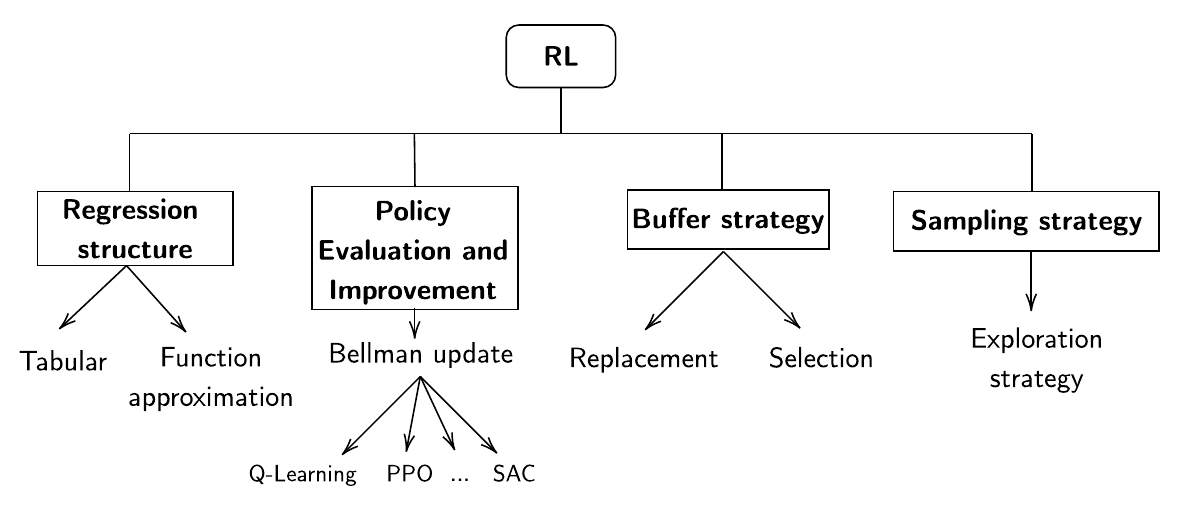}
    \caption{Reinforcement learning: design choices}
    \label{fig:rl}
\end{figure}

Each part of the algorithm can be instantiated in various ways, constituting the design choices of RL, as illustrated in Figure~\ref{fig:rl}. The following sections discuss the role of each part and give a few examples of possible instantiations.

\begin{algorithm}[tb]
\caption{Reinforcement learning process.\label{algo:batched_td}}
\textbf{Input}: Stopping criterion $stop$, Environment $env$.\\
\textbf{Output}: An optimized policy $\pi$.
\begin{algorithmic}[1]
\State $\pi  = \mathit{Initialize()}$
\State $\mathbb B = \emptyset$

\While{$\neg \mathit{stop}$}
    \State $experiences = \mathit{Sample}(env, \pi)$ 
    \State $\mathbb B = \mathit{UpdateBuffer}(\mathbb B, experiences)$
    \State $\Tilde{v}^\pi =  \mathit{EvaluatePolicy}(\pi)$
    \State $\pi = \mathit{ImprovePolicy}(\pi, \Tilde{v}^\pi, \mathbb B)$
\EndWhile
\State \textbf{return }$\pi$
\end{algorithmic}
\end{algorithm}

\subsection{Regression Structure}

A cornerstone design point in RL regards how to encode the policy function.  Early RL algorithms often represented the policy using a tabular format. For example, Q-Learning \cite{watkins_q-learning_1992} stores the action-value estimates, $\Tilde{q}(s, a)$ in such a table. From this structure, a policy can be derived by selecting the action with the highest q-value from each state, \textit{i.e.} $\pi(s, a) = \argmax_{a \in \mathbb A} \Tilde{q}(s, a)$. This is known as a greedy deterministic policy, as it always chooses the action with the largest expected reward.

However, this kind of approach does not scale to highly dimensional problems, \textit{e.g.} with continuous states or actions. Hence, recent algorithms use function approximation based on regression, such as trees or deep neural networks (DNNs) \cite{mnih_human-level_2015}. In such settings, with the policy $\pi_{\theta}$ being parameterized by a set of parameters $\theta \in \Theta$ ($\Theta$ being the parameter space), the RL problem boils down to finding an optimal assignment of parameters $\theta^* = \argmax_{\theta \in \Theta} v^{\pi_\theta}$. Finally, recent algorithms often rely on the use of multiple regression structures. For example, in actor-critic settings, one structure, the actor, aims at representing the probability of taking an action (\textit{i.e.} the policy), while another structure, the critic, estimates the action values \cite{sutton_reinforcement_2018}.

\subsection{Policy Evaluation and Improvement}

RL algorithms usually rely on estimated policy evaluations to bootstrap their policy improvement process~\cite{sutton_reinforcement_2018}. For example, from an experience tuple $(s_t, a_t, r_t, s_{t+1})$ sampled from the environment and its current estimations $\Tilde{q}(s,a)$, the well-known Q-Learning algorithm \cite{watkins_q-learning_1992} updates its estimations using the following relation:
\begin{equation}
\label{eq:q-learn}
\Tilde{q}(s_t,a_t) \leftarrow \Tilde{q}(s_t,a_t) + \alpha\left(r_t + \gamma \max_{a'\in \mathbb A} \Tilde{q}(s_{t+1},a') - \Tilde{q}(s_t,a_t)\right)
\end{equation}
\noindent where $\alpha \in [0, 1]$ is the learning rate, and the term $r_t + \gamma \max_{a'\in \mathbb A} \Tilde{q}(s_{t+1},a') - \Tilde{q}(s_t,a_t)$ is called temporal difference error (TD-error). Many other update relations have been published over the years: policy gradient methods such as REINFORCE \cite{sutton_reinforcement_2018}, or actor-critic approaches such as PPO \shortcite{schulman_proximal_2017} and SAC \shortcite{haarnoja_soft_2018}. 

\subsection{Buffer Strategy}

In recent algorithms, policy updates are often batched using an experience buffer. This allows learning multiple times from an experience by replaying it, but also speeding up the policy improvement steps by performing mini-batch gradient descent in deep RL. Two choices linked to buffers arise: deciding which experiences in the buffer will be replaced by new ones, and which experiences to select to update the policy.

\paragraph{Replacement.} While the simplest replacement criterion is based on recency, more elaborate techniques have also been studied. For example, \shortciteA{de_bruin_improved_2016} propose using two experience buffers: one that is close to the current policy (recency criterion), while another one keeps the stored experiences close to a uniform distribution over the state-action space. This allows the computation of more robust policies and reduces the need for continuous, thorough exploration.

\paragraph{Selection.} The most straightforward method for selecting experiences from the buffer is to choose them uniformly. However, research has shown in various articles that more intelligent experience selection strategies can significantly improve results in practice. For instance, prioritized sampling, as discussed in \shortciteA{schaul_prioritized_2016}, is one such approach that has been shown to yield superior outcomes.

\subsection{Sampling Strategy}
The performance of an RL agent depends on which experiences were collected in the environment to learn its policy. Thus, the question of which action to choose in each state is crucial in such algorithms. To reach good performances, an agent must ideally sample (1) multiple times the same state-action pairs so as to compute good estimates of the environment dynamics (which can be stochastic), (2) regions leading to good rewards, to obtain good performances, and (3) unexplored regions, to find potentially better regions. The tension between points (2) and (3) is commonly referred to as the \textit{exploration-exploitation dilemma}. Typically, the agent faces the dilemma of strictly adhering to its current policy (exploitation) and making exploratory moves guided by specific criteria. A widely used approach in this scenario is known as $\epsilon$-greedy, which introduces an element of randomness to facilitate exploration within the environment. Additionally, certain methods suggest building a model, such as a map, of the environment. This model provides the agent with a broader perspective, enabling more systematic exploration strategies. This concept is referred to as ``state-based exploration" in the work of \shortciteA{moerland_model-based_2023}.

\subsection{Summary}

This section presented RL, which allows the automated learning of agent's behaviors. The whole learning process is based on reinforcement towards good signals, provided by the reward function. However, in MDPs, these reward functions are limited to scalar signals, which limits the applicability of such techniques. Indeed, real-world scenarios often involve making compromises between multiple objectives, as noted in~\shortciteA{vamplew_scalar_2022}. To initiate the discussion towards more complex reward schemes, the following section presents the world of multi-objective optimization through the prism of decomposition techniques.

\section{Multi-Objective Optimization Based on Decomposition}
\label{sec:MOO/D}

Multi-objective optimization aims at optimizing problems involving multiple conflicting objectives. In such settings, a solution $x$ is evaluated using a $m$-dimensional vector function, where $m$ is the number of objectives. Formally, the optimization problem can be expressed as $\max \vec f(x) = \max (f_1(x), ..., f_m(x))$ subject to $x \in \Omega$, where $\Omega$ is the decision space (\textit{i.e.} search space), and $\vec f$ is the objective function. 

\paragraph{Solution concepts.} In cases where the decision maker (DM) has known preferences over objectives \textit{a priori}, these problems can be simplified to single-objective problems by converting the objective values into scalars using a scalarization function $g(\vec f(x)): \mathbb R^m \rightarrow \mathbb R$. However, many approaches and real-life scenarios cannot make this assumption and instead operate within the \textit{a posteriori} setting, where the DM's preferences are not known in advance.

In the \textit{a posteriori} setting, where the evaluation of each solution maintains a vector shape, algorithms commonly rely on the concept of Pareto dominance to establish an order among solutions. A solution $x$ is said to Pareto dominate another solution $x'$ if and only if it is strictly better for at least one objective, without being worse in any other objective. Formally:

\begin{equation*}
    x \succ_P x' \iff (\forall i : f_i(x) \geq f_i(x')) \wedge 
    (\exists j : f_j(x) > f_j(x')).
\end{equation*}

This definition does not impose a total ordering on all solutions. For instance, it is possible that two solutions' evaluations, such as $(1,0)$ and $(0,1)$, may not dominate each other. These solutions are referred to as \textit{Pareto optimal}, signifying that they are both potentially optimal solutions as long as the DM's preferences remain unknown. Consequently, the primary goal of the optimization process is to identify a collection of Pareto optimal solutions, referred to as the \textit{Pareto set}. The evaluations associated with these solutions constitute the \textit{Pareto front} (PF). Upon receiving a set of solutions, the DM can then make an informed choice, taking into account the trade-offs presented in the PF. Formally, a PF is defined as follows:

\begin{equation}
\label{eq:pf}
     \mathcal{F} \equiv \{\vec f(x) \ |\ \nexists \ x' \: \mathrm{ 
 s.t. } \ x' \succ_P x \}.
\end{equation}

\begin{figure}
    \centering
    \includegraphics[width=0.7\textwidth]{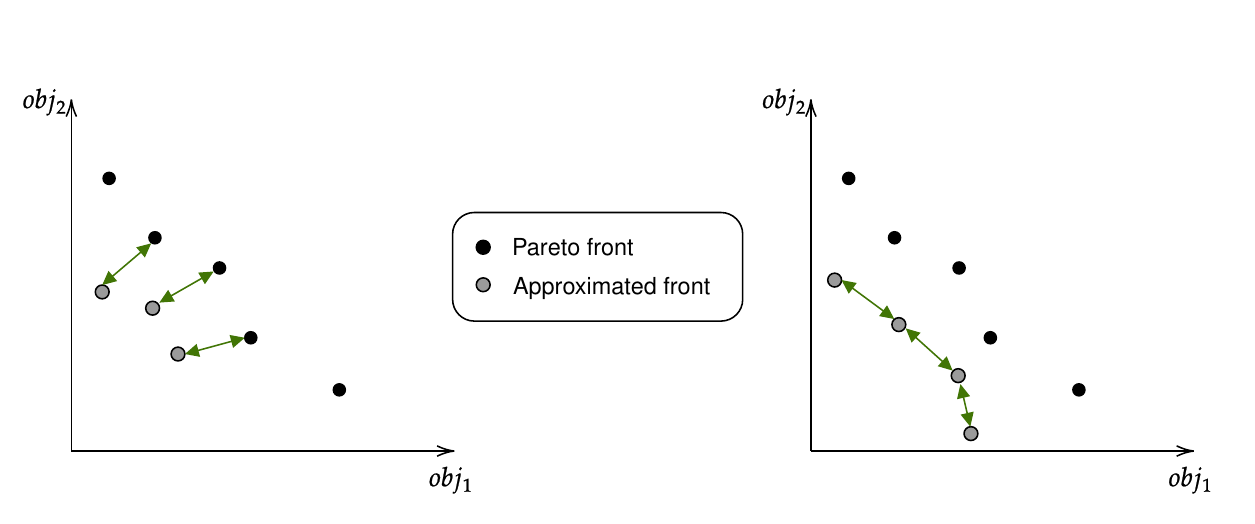}
    \caption{Illustration of Pareto front approximations. In the left part, the convergence aspect of the approximated front is represented by the arrows. In the right part, the diversity aspect is represented by the arrows.}
    \label{fig:convergence_diversity}
\end{figure}

\paragraph{Approximated optimization methods.} In most scenarios, real-world problems are often too hard to solve using exact methods. Hence, algorithms often provide an approximation of the Pareto set (and its corresponding front) by relying on metaheuristics. A good approximation of the PF is characterized by two criteria: (1) \textit{convergence} with the true solution, to present solutions of good quality to the DM, and (2) \textit{diversity}, to present a wide range of compromises to the DM. Examples of Pareto fronts showing the importance of both criteria are shown in Figure~\ref{fig:convergence_diversity}.

\paragraph{Single solution and population-based methods.} The literature of \textit{a posteriori} MOO is generally divided into two classes of algorithms: \textit{single solution based} and \textit{population-based} \cite{talbi_metaheuristics_2009}. The first class maintains and improves a single solution at a time and loops through various preferences, trajectories, or constraints to discover multiple solutions on the PF, \textit{e.g.} \shortciteA{czyzzak_pareto_1998,hansen_tabu_2000}. The second class maintains a set of solutions, called \textit{population}, that are jointly improved over the search process, \textit{e.g.} \citeA{zhang_moead_2007,alaya_ant_2007}.  Notably, this population-based approach is currently considered the state of the art for solving MOPs due to the performance of these algorithms. Consequently, this work will concentrate on population-based algorithms.

\begin{figure}
    \centering
    \includegraphics[width=0.9\textwidth]{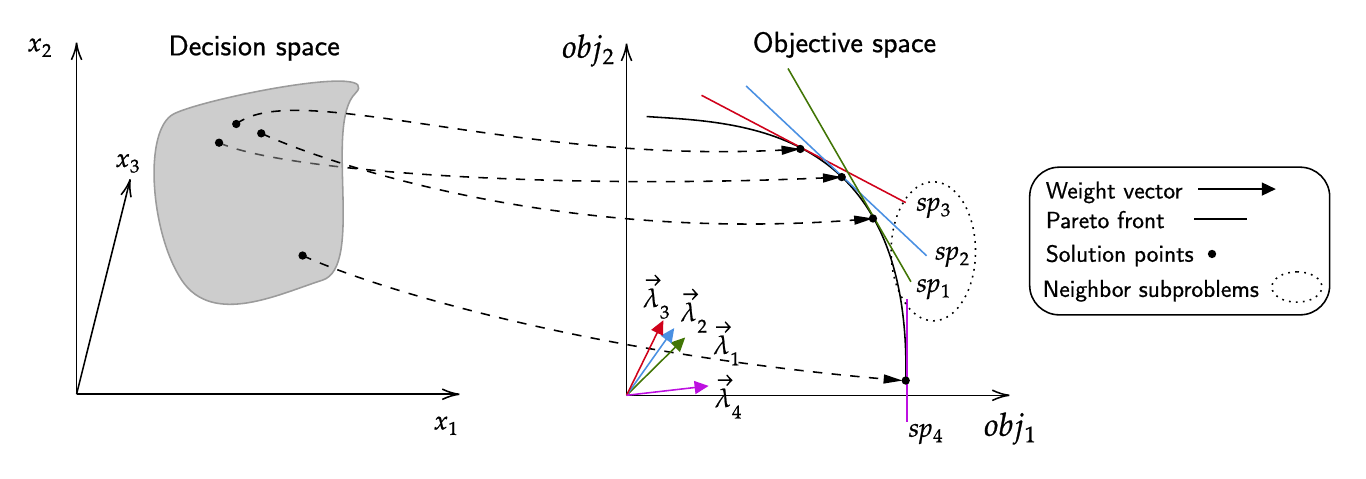}
    \caption{The decomposition in the objective space idea: split the multi-objective problem into various single-objective problems $sp_n$ by relying on a scalarization function (weighted sum in this case). $sp_1$, $sp_2$, and $sp_3$ are considered to be neighbors since their associated weight vectors are close to each other while $sp_4$ is not considered to be in the neighborhood.}
    \label{fig:decomposition}
\end{figure}

\paragraph{Decomposition.} Multi-objective optimization based on decomposition (MOO/D) relies on the fundamental concept of dividing a MOP into several SOPs using a scalarization function. This function, represented by $g: \mathbb R^m \rightarrow \mathbb R$, employs associated weights represented by the vector $\vec\lambda$. This methodology, as depicted in Figure~\ref{fig:decomposition}, facilitates the approximation of the PF by solving the SOPs corresponding to different weight vectors. Each of these weight vectors is designed to target specific regions of the PF, providing a comprehensive exploration of the objective space. Moreover, decomposition usually simplifies the problem and offers a simple way to parallelize the search process. This technique is applicable in both the context of \textit{single-solution based} and \textit{population-based} algorithms.

In the population-based setting, the assumption that neighboring subproblems share common solution components is often utilized, enabling these subproblems to cooperate by exchanging information. This cooperation typically improves the optimization process.

\begin{algorithm}[tb]
\caption{Population based MOO/D, high-level framework. \label{algo:MOO/D}}
\textbf{Input}: Stopping criterion $stop$, Scalarization method $g$, Exchange trigger $exch$.\\
\textbf{Output}: The approximation of the Pareto set stored in the external archive population $\mathbb{EP}$.
\begin{algorithmic}[1]
\State $\mathbb{P}, \mathbb W, \mathbb Z  = \mathit{Initialize()}$
\State $\mathbb{EP} = \mathit{Prune(}\mathbb P)$
\State $\mathbb N = \mathit{InitializeNeighborhood(\mathbb P, \mathbb W)}$

\While{$\neg \mathit{stop}$}
    \State $\mathit{individuals}, \mathbb W', \mathbb Z' = \mathit{Select(}\mathbb P, \mathbb W, \mathbb Z)$
    \State $\mathit{candidates} = \mathit{Search(individuals}, \mathbb W', \mathbb Z', g, \mathit{exch})$
    \State $\mathbb P = \mathbb P \cup \mathit{candidates}$
    \State $\mathit{evaluations} = \forall c \in \mathit{candidates}: \vec f(\mathit{c}) $ 
    \State $\mathbb{EP} = \mathit{Prune(}\mathbb{EP} \cup \mathit{candidates}, \mathit{evaluations})$
    \State $\mathbb W, \mathbb Z = \mathit{Adapt(}\mathbb W, \mathbb Z, \mathbb{EP})$
    \State $\mathbb N = \mathit{UpdateNeighborhood(} \mathbb P, \mathbb N, \mathbb W)$
    \State $\mathit{Cooperate(}\mathbb N)$
\EndWhile
\State \textbf{return }$\mathbb{EP}$
\end{algorithmic}
\end{algorithm}

A generic framework based on such decomposition techniques can be found in Algorithm~\ref{algo:MOO/D}. MOO/D maintains a population of solutions $\mathbb P$, a set of weights $\mathbb W$, and reference points $\mathbb Z$ to apply in the scalarization function $g$. The best individuals are kept in an external archive population $\mathbb{EP}$ during the optimization process according to a criterion defined by the $Prune$ function, \textit{e.g.} Pareto dominance (Equation~\ref{eq:pf}). At each iteration, a set of individuals from the population, along with weights ($\mathbb{W'} \subseteq \mathbb W$) and reference points ($\mathbb{Z'} \subseteq \mathbb{Z}$) are selected as starting points to search for better solutions until an exchange criterion ($exch$) is triggered (lines 5--6). Then, the generated candidates are integrated into the population and archive based on their evaluated performance (lines 7--9). Next, the algorithm adapts the weights and reference points according to the current state of the PF (lines 10--11). Additionally, some knowledge is exchanged between subproblems in the same neighborhood (line 12). Finally, the algorithm returns the set of non-dominated solutions seen so far (line 14). 

\begin{figure}
    \centering
    \includegraphics[width=0.99\textwidth]{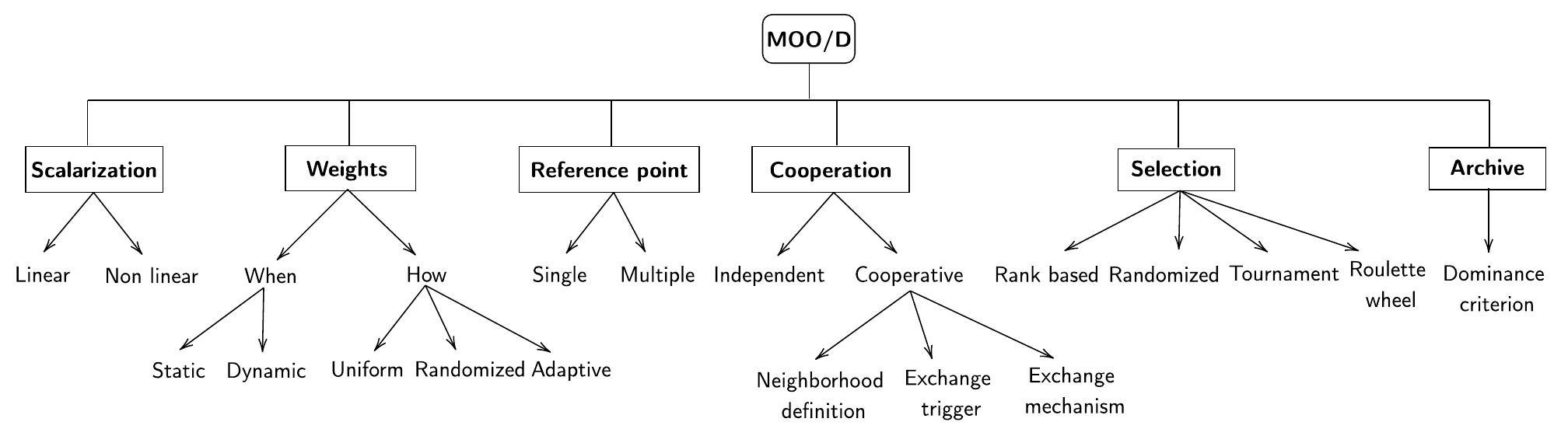}
    \caption{Design choices of multi-objective optimization based on decomposition (MOO/D).}
    \label{fig:moo_d}
\end{figure}

Over the years, many variations of the MOO/D framework have been proposed, and recurring patterns were identified, \textit{e.g.} Figure~\ref{fig:moo_d} has been compiled from various sources \shortcite{talbi_metaheuristics_2009,santiago_survey_2014,xu_survey_2020}. The rest of this chapter explains each of the building blocks that can be instantiated in different ways in such a framework. Some illustrative examples and notable articles are also pointed out in the discussion. For further references, some surveys dedicated to decomposition techniques in MOO can be found in \shortciteA{santiago_survey_2014,xu_survey_2020}.

\subsection{Scalarization Functions}
Various scalarization functions $g$ have been the subject of studies over the last decades. These allow the MOP to be decomposed into multiple simpler SOPs. Moreover, scalarization functions are usually parameterized by weight vectors that target different areas of the objective space. 

\paragraph{Linear scalarization.} The most common and straightforward scalarization technique is the weighted sum: $g^{\text{ws}}(x) = \sum_{i=1}^m \lambda_i f_i(x) = \vec\lambda^\intercal \vec f(x)$. This method is easy to comprehend and enables the specification of weights as percentages to express preferences between the objectives. However, this approach has limitations. With this kind of simplification, the subproblems become linear, which means they cannot accurately capture points in the concave region of the PF \cite{marler_weighted_2010}.

\paragraph{Non-linear scalarization.} In response to the limitations of linear scalarization, alternative non-linear techniques, such as the Chebyshev scalarization, have been investigated. This scalarization function, also known as the norm $L^\infty$, is defined as the maximum weighted distance to a utopian reference point $\vec z$, expressed as $g^{\text{ch}}(x) = \max_{i \in [1, m]} |\lambda_i (f_i(x) - z_i)|$. Note that in this particular case, the goal is to minimize the scalarized values instead of maximizing like in the linear case. Using such non-linear techniques enables the identification of points within the concave regions of the PF. See for example the work of~\citeA{emmerich_tutorial_2018} for some visual examples. Intriguingly, some studies, such as \shortciteA{ishibuchi_simultaneous_2010}, have also proposed the combination or adaptation of various scalarization techniques to leverage their respective advantages.

\subsection{Weights}
Scalarization functions rely on weight vectors $\vec\lambda \in \mathbb W$ to target different points in the PF. Therefore, the way these weights are generated is crucial to obtain good solutions. There are two design choices for the weights: (1) when and (2) how to (re-)generate them. 

\paragraph{When to generate weights?} The simplest approach is to fix the weights before any search is started (\textbf{static}). However, the shape of the Pareto front, being unknown at that time, can be complex and may require adapting the weights during the search to focus on sparse areas, \textit{i.e.}~where the Pareto front is not well estimated. Thus, multiple ways to adapt the weights during the search have been published, \textit{e.g.} \shortciteA{qi_moead_2013}. This is referred to as \textbf{dynamic} weights in this work.

\paragraph{How to generate weights?} Weights can be assigned through different approaches, such as \textbf{uniform} distribution across the objective space \shortcite{das_normal-boundary_2000,blank_generating_2021}, \textbf{randomized} processes, or \textbf{adaptation} based on evolving knowledge during the search \cite{qi_moead_2013,czyzzak_pareto_1998}. Weight adaptation strategies can vary, including focusing on underrepresented regions in the estimated PF or predicting potential improvements in the objective space~\shortcite{dubois-lacoste_improving_2011}. Pareto simulated annealing (PSA)~\shortcite{czyzzak_pareto_1998}, for instance, exemplifies this adaptation approach, as it adjusts an individual's weights in response to its current evaluation and proximity to non-dominated solutions. Formally, for an individual $x$ and its nearest non-dominated neighbor $x'$, PSA modifies the weights attached to the SOP leading to $x$ using:

\begin{equation}
\label{eq:PSA}
  \lambda^x_j =
    \begin{cases}
      \delta \lambda^x_j & \text{if $f_j(x) \geq f_j(x')$}\\
      \lambda^x_j/\delta & \text{if $f_j(x) < f_j(x')$},
    \end{cases}       
\end{equation}

\noindent with $\delta$ being a constant close to 1, typically $\delta=1.05$. This mechanism, encapsulated within the \textit{Adapt} function of MOO/D, allows the fine-tuning of SOPs to excel at objectives where they already exhibit proficiency. This, in turn, encourages the SOPs to move away from their neighboring solutions, ultimately enhancing the diversity within the estimated PF.

\subsection{Reference Points}
Certain scalarization techniques, such as Chebyshev, depend on the selection of an optimistic or pessimistic point in the objective space to serve as a reference point $\vec z$. Similarly to weight vectors, the choice of these reference points has a significant influence on the final performance of the MOO algorithm. The careful determination of these reference points is crucial, as it profoundly impacts the algorithm's ability to approximate the Pareto front effectively. Hence, multiple settings for reference points exist.

\paragraph{Single reference point.} A straightforward approach to setting the reference point is to fix it before commencing the optimization process \shortcite{zhang_moead_2007}. In this manner, the reference point remains constant throughout the optimization, allowing for a controlled and deterministic approach.

\paragraph{Multiple reference points.} Nevertheless, setting the reference point beforehand can sometimes result in suboptimal outcomes for the algorithm. For this reason, some studies suggest dynamic adaptation of the reference point during the search, as exemplified by the work of \shortciteA{liu_adapting_2020}. This adaptive approach enables the reference point to evolve and align with the changing characteristics of the Pareto front, potentially improving the quality of the results.

\subsection{Cooperation}
The most straightforward approach to solving a MOP using decomposition is to \textbf{independently} address all generated SOPs. This concept underpins the first single-solution decomposition-based algorithms, such as the one introduced by \citeA{czyzzak_pareto_1998}. Subsequently, the idea of promoting \textbf{cooperation} among subproblems in close proximity (referred to as neighbors) emerged \cite{zhang_moead_2007}. Numerous studies have indicated that cooperation among these subproblems can accelerate the search process and yield superior performance. Within this cooperative framework, several design choices come into play: (1) the definition of neighborhood relationships, (2) the timing of information exchange, and (3) the nature of the knowledge to be shared. 

\paragraph{Neighborhood definition.} A neighborhood $\mathbb N$ can be constituted of zero (no cooperation), multiple, or all other subproblems. The most common way to define neighborhoods between SOPs is to rely on the Euclidean distance of their associated weight vectors; see Figure~\ref{fig:decomposition}. Yet, other techniques have also been published such as \shortciteA{murata_specification_2001}, which relies on the Manhattan distance between weight vectors instead.

\paragraph{Exchange trigger.} Information exchange between SOPs can occur at various moments throughout the search process, with the timing often determined by an exchange trigger ($exch$ in Algorithm~\ref{algo:MOO/D}). There exist three primary approaches to information exchange. \textbf{Periodic exchange} involves regular and predetermined information sharing, such as at every iteration. \textbf{Adaptive exchange}, on the other hand, responds dynamically to events occurring in the search process, such as when specific improvements are achieved. Finally, \textbf{continuous exchange} is usually achieved through a shared structure that constantly disseminates information to guide the search process for all neighboring SOPs. 

\paragraph{Exchange mechanism.} The method of cooperation, denoted as $\mathit{Cooperate}$ in MOO/D, among subproblems is closely associated with the specific search algorithm being employed. Presently, many decomposition techniques are integrated with genetic algorithms, which perform information exchange among SOPs by employing crossover operators on the solutions. This enables the sharing of solution components and relevant data. An alternative approach involves sharing part of the search memory, as seen in the case of ant colony algorithms, where elements like the pheromone matrix are shared among subproblems to guide the search process~\cite{ke_moead-aco_2013}. 

\subsection{Selection}
As the weight vectors and reference points attached to individuals within the population $\mathbb P$ can change dynamically, an exponential number of combinations arises for each optimization step. To address this, various selection mechanisms (denoted as \textit{Select} in MOO/D) have been developed to choose a subset of individuals, along with their associated weights and reference points, for each search iteration.

For example, existing approaches \textbf{rank} solutions according to their scalarized values and select the best ones observed thus far, using a static weight vector and reference point~\cite{zhang_moead_2007}. In contrast, another method involves random generation of new weight vectors, followed by \textbf{tournament-style selection} to determine which solution will initiate the local search phase~\cite{ishibuchi_implementation_2004}. Alternative techniques propose \textbf{random selection} or systematic iteration through the available choices, akin to a \textbf{roulette wheel} selection process~\cite{talbi_metaheuristics_2009}.

\subsection{Archive}
With the dynamic parts of the algorithms explained above, a SOP search may find a worse solution after adaptation. This could lead to a degradation of the quality of the population. To solve this issue, modern algorithms often rely on the concept of external archive population ($\mathbb{EP}$), which stores the non-dominated solutions seen so far. The Pareto dominance criterion (Equation~\ref{eq:pf}) can be used as a pruning function ($\mathit{Prune}$ in MOO/D) to determine which individuals to keep in the archive. In case there are too many incomparable solutions, the size of the archive can be limited by using additional techniques, \textit{e.g.} crowding distance \shortcite{deb_fast_2002}. 

\subsection{Summary}
Section~\ref{sec:RL} presented the RL problem, as well as its building blocks, and concluded by discussing its limited expressivity for multi-objective problems. This section presented an overview of MOO/D, which aims at solving multi-objective problems. However, there are a few notable differences between MOO/D and RL: (1) in MOO, the dynamics of the environment ($\vec f$, and the MOP model) are known by the algorithm and are generally deterministic, whereas RL involves solving sequential problems with unknown and potentially stochastic dynamics, (2) in MOO, the solutions are assignments of the decision variables of the problem, whereas they are reusable functions in RL.\footnote{Generating reusable functions to optimize problems is known as Hyper-Heuristics in optimization \shortcite{burke_hyper-heuristics_2013}.}

The upcoming section introduces the core contribution of this work, multi-objective reinforcement learning based on Decomposition (MORL/D). This approach establishes a taxonomy that bridges the gap between the two previously discussed fields, creating a shared vocabulary and facilitating the precise identification of research contributions within the context of MORL. Furthermore, this taxonomy enables researchers to recognize relevant concepts and findings from other related disciplines, fostering cross-disciplinary knowledge exchange. Moreover, the section reveals a unified framework that aligns with this taxonomy, demonstrating how techniques from both the realms of RL and MOO/D can be integrated. 

\section{Multi-Objective Reinforcement Learning Based on Decomposition (MORL/D)}
\label{sec:morld}

To extend RL to multi-objective problems, MORL models the problem as a multi-objective MDP (MOMDP). A MOMDP alters the MDP by replacing the reward function with a vectorial reward signal $\vec r: \mathbb S \times \mathbb A \times \mathbb S \rightarrow \mathbb R^m$ \cite{roijers_multi-objective_2017}. As mentioned earlier, one of the key differences between MOO and RL lies in the fact that RL aims at learning (optimizing) a policy, while MOO aims at optimizing a Pareto set of solutions. In the middle ground, MORL aims at learning a Pareto set of policies $\Pi$, with each policy parameterized by $\theta \in \Theta$ under function approximation. 

\begin{figure}
    \centering
    \includegraphics[width=0.9\textwidth]{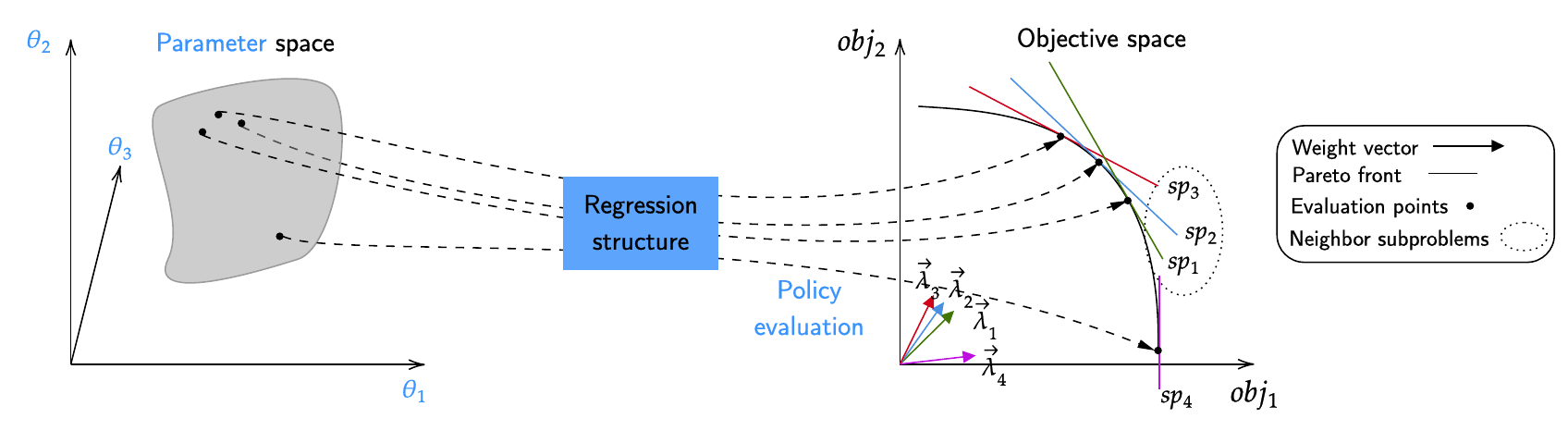}
    \caption{The decomposition idea applied to MORL. Blue parts emphasize the parts coming from RL, while black parts come from MOO. The optimization is looking for the best parameters for the regression structure to generate good policies. The idea of neighbor policies is that policies that have similar parameters should lead to close evaluations.}
    \label{fig:decomposition_RL}
\end{figure}

In MORL, as in classical RL, the evaluation of policies is based on a sequence of decisions, and one usually runs a policy $\pi$ for a finite amount of time to estimate its vector value function $\vec{v^\pi}$. This value, which can be considered as the equivalent of $\vec f(x)$ for MORL (see Figure~\ref{fig:decomposition_RL}), is the direct vectorial adaptation of $v^\pi$. Formally, a MORL policy has a vector value that is computed using the following:

\begin{equation}
\vec{v^\pi}  = \Ex\left[\sum_{t=0}^\infty \gamma^t \vec r(s_t, a_t, s_{t+1}) | \pi, \mu_0\right].
\end{equation} 


\noindent Similar to MOO, in MORL, the DM's preferences are typically unknown during the training phase. This lack of knowledge prevents the establishment of a total ordering of actions at training time, as these actions are often Pareto incomparable. As a result, MORL algorithms are typically designed to be trained offline and executed once a specific trade-off has been chosen~\cite{hayes_practical_2022}. Our work focuses primarily on the MORL learning phase, which involves the process of training multiple policies to present a PF to the user. Notably, multiple MORL algorithms, environments, and software tools aiming at tackling such an issue have recently been published~\shortcite{roijers_survey_2013,hayes_practical_2022,felten_toolkit_2023}.

\paragraph{Pareto-based methods.} In the realm of MORL, certain algorithms adopt an approach that involves directly learning a set of policies within the regression structure~\cite{van_moffaert_multi-objective_2014,ruiz-montiel_temporal_2017}. In these algorithms, each Q-value is designed to store a set of Pareto optimal value vectors that can be achieved from the current state. However, it is worth noting that these algorithms are currently limited to tabular structures and face challenges when applied to problems with high dimensions. Despite efforts to develop Pareto-based MORL algorithms using function approximation techniques, learning to produce sets of varying sizes for different state-action pairs remains a challenging problem~\cite{reymond_pareto-dqn_2019}. Furthermore, even when the agent can learn a set of optimal policies for each action, questions persist about how to effectively order actions during the training phase as these are Pareto incomparable~\shortcite{felten_metaheuristics-based_2022}.

\paragraph{Decomposition-based methods.} To solve such issues, multiple MORL algorithms relying on decomposition (MORL/D) techniques have been presented~\shortcite{felten_morld_2022}. The utilization of decomposition techniques offers several advantages for MORL. Firstly, by scalarizing rewards with different weights, these algorithms can often leverage single-objective RL techniques to learn multiple optimal policies. This allows MORL to directly benefit from advancements in RL research. Additionally, scalarization provides a method for ordering the evaluations of potentially Pareto incomparable actions during the training phase, enabling the selection of actions in a greedy manner based on certain weights. Based on these findings, it seems natural that part of the techniques from MOO/D can be transferred to MORL/D. 

In this trend, in the same vein as single solution-based MOO, the most straightforward algorithm that we call vanilla outer loop, proposes to sequentially train single-objective RL with different weights applied in the scalarization to target different parts of the objective space \cite{roijers_multi-objective_2017}. Obviously, when compared to single-objective RL, this naive approach requires significantly more samples from the environment to compute various policies. Although MORL is a relatively new field of research, various enhancements and variations, often based on cooperation schemes, have already been proposed to enhance the sample efficiency of MORL/D algorithms (Figure~\ref{fig:decomposition_RL}). Nevertheless, to the best of our knowledge, no prior work has undertaken the task of systematically studying these recurring patterns and categorizing them within a comprehensive taxonomy, a gap that this present work seeks to fill.


\subsection{Taxonomy and Framework}

\begin{algorithm}[tb]
\caption{MORL/D high-level framework. Blue parts emphasize concepts coming from RL while black parts come from MOO/D.\label{algo:MORL/D}}
\textbf{Input}: Stopping criterion $stop$, Population size $n$, Scalarization method $g$, Exchange trigger $exch$, Environment $env$, Update passes $u$.\\
\textbf{Output}: The approximated Pareto set stored in the external archive population $\mathbb{EP}$.
\begin{algorithmic}[1]
\State $\textcolor{myblue}{\Pi}, \mathbb W, \mathbb Z  = \mathit{Initialize(n)}$
\State $\textcolor{myblue}{evals = EvaluatePolicies(\Pi)}$ 
\State $\mathbb{EP} = \mathit{Prune(}\textcolor{myblue}{\Pi, evals})$
\State $\mathbb N = \mathit{InitializeNeighborhood(}\textcolor{myblue}{\Pi}, \mathbb W)$
\State $\textcolor{myblue}{\mathbb B = \emptyset}$

\While{$\neg \mathit{stop}$}
    \State $\textcolor{myblue}{\pi}, \vec\lambda, \vec z = \mathit{Select(}\textcolor{myblue}{\Pi}, \mathbb W, \mathbb Z)$
    \State $\textcolor{myblue}{\mathit{experiences} = \mathit{Sample(}env, \pi}, g, \vec\lambda, \vec z, exch\textcolor{myblue}{)}$
    \State $\textcolor{myblue}{\mathbb B = \mathit{UpdateBuffer(}\mathbb B, \mathit{experiences})}$
    \State $\textcolor{myblue}{\mathit{\Pi} = \mathit{ImprovePolicies(\Pi, }\: g, \mathbb W, \mathbb B, u)}$
    \State $\textcolor{myblue}{evals = EvaluatePolicies(\Pi)}$
    \State $\mathbb{EP} = \mathit{Prune(}\mathbb{EP} \cup \textcolor{myblue}{\Pi}, evals)$
    \State $\mathbb W, \mathbb Z = \mathit{Adapt(}\mathbb W, \mathbb Z, \mathbb{EP})$
    \State $\mathbb N = \mathit{UpdateNeighborhood(} \textcolor{myblue}{\Pi}, \mathbb N, \mathbb W)$
    \State $\mathit{Cooperate(}\mathbb N)$
\EndWhile
\State \textbf{return }$\mathbb{EP}$
\end{algorithmic}
\end{algorithm}

Algorithm~\ref{algo:MORL/D} defines a high-level skeleton of the MORL/D framework, a middle-ground technique between MOO/D and RL aimed at finding a set of solutions to MOMDPs. Here, the population of policies (solutions) is denoted $\Pi$ to clearly identify the different solution concepts with RL and MOO. The algorithm starts by initializing the policies, weights, reference points, Pareto archive, neighborhoods, and buffer (lines 1--5). At each iteration, the algorithm samples some experiences from the environment by following a chosen policy (lines 7--8). These experiences are then included in an experience buffer and used to improve the policies by sampling $u$ batches from the buffer for each policy in the population (lines 9--10). After improvement, the policies are evaluated, and the Pareto optimal policies are included in the Pareto archive (lines 11--12). Then, the weights, reference points, and neighborhood are adapted (lines 13--14), and subproblems can cooperate with each other (line 15). Finally, the set of non-dominated policies is returned (line 17). 

\begin{figure}
    \centering
    \includegraphics[width=0.99\textwidth]{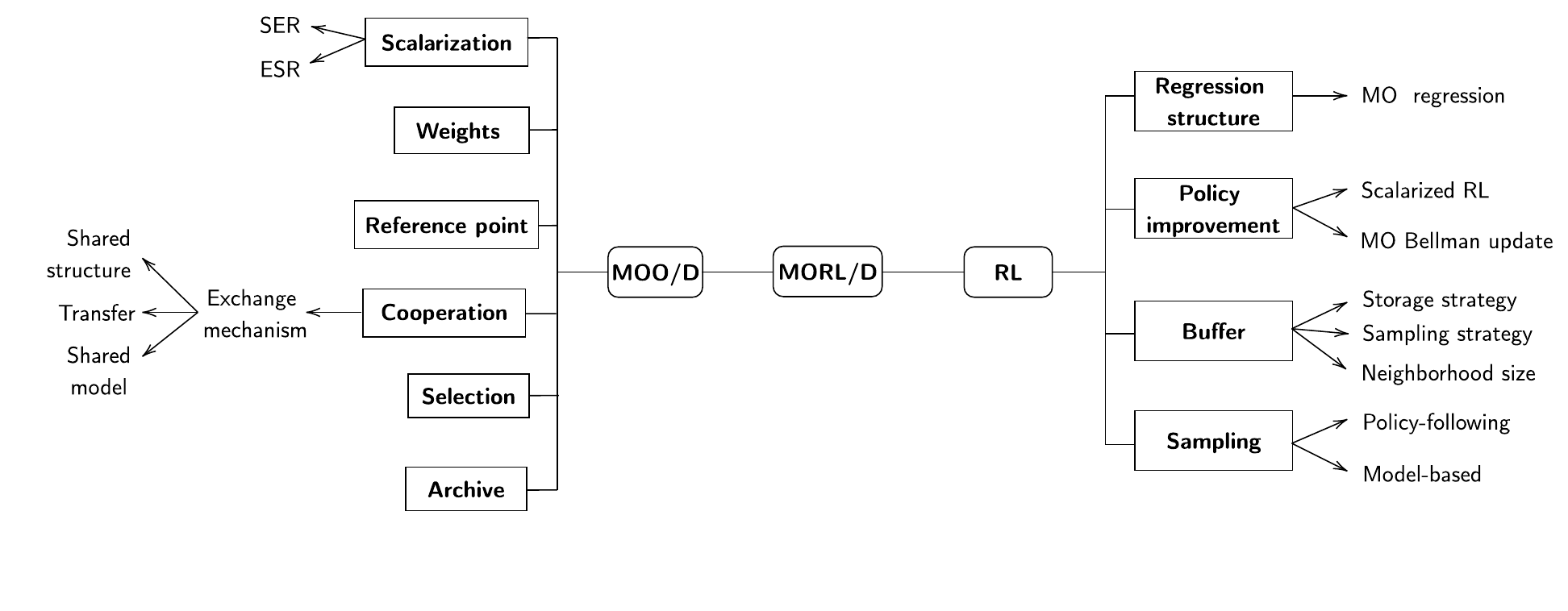}
    \caption{The Multi-Objective Reinforcement Learning based on Decomposition (MORL/D) taxonomy. Some traits from MOO/D and RL are directly applicable to this technique. Yet some alterations tailored for MORL have been published and are expanded in the figure.}
    \label{fig:morld}
\end{figure}

Naturally, MORL/D benefits from a rich pool of techniques inherited from both MOO/D and RL. Some elements can be directly transferred and instantiated, while others require specific adaptations and novel approaches. Figure~\ref{fig:morld} visually highlights these design choices, and the following parts delve into some of these choices in more depth, accompanied by examples from existing MORL literature that employ these techniques.

\subsection{Common Design Choices with MOO/D}
Most of the building blocks previously identified in MOO/D can immediately be applied to MORL/D. Indeed, weight vectors or reference points generation and adaptation schemes, Pareto archive, and individual selection mechanisms can be applied straightforwardly. However, some building blocks require particular attention; these are discussed in more detail below.

\subsubsection{Scalarization}
\label{sec:scalarization_morld}

In RL, the evaluation of a policy comes from multiple rewards, collected over a sequence of decisions made by the agent. To reduce these to a scalar for evaluation, RL usually relies on the expectation of the discounted sum of rewards. In MORL/D, to have a total ordering of Pareto incomparable actions or to rely on single-objective policy improvements, a scalarization function is used. To transform this sequence of reward vectors into scalars for evaluation in MORL, the scalarization function can be applied before or after the expectation operator. Hence, the algorithm can optimize for the Expected Scalarized Return (\textbf{ESR}), or the Scalarized Expected Return (\textbf{SER}). 

When using a linear scalarization, both settings are equivalent. However, the ESR and SER settings lead to different optimal policies when the scalarization is not linear \cite{roijers_multi-objective_2018}. Formally:

\begin{equation*}
    \pi^*_{\text{SER}} = \argmax_\pi g \left( \mathop{\mathbb{E}}\bigg[\sum_{t=0}^{\infty} \gamma^t \vec r_t | \pi, s_0\bigg] \right) \neq  \argmax_\pi \mathop{\mathbb{E}}\left[g\bigg(\sum_{t=0}^{\infty} \gamma^t \vec r_t\bigg) | \pi, s_0\right] = \pi^*_{\text{ESR}}.
\end{equation*}

\begin{figure}
    \centering
    \includegraphics[width=0.4\textwidth]{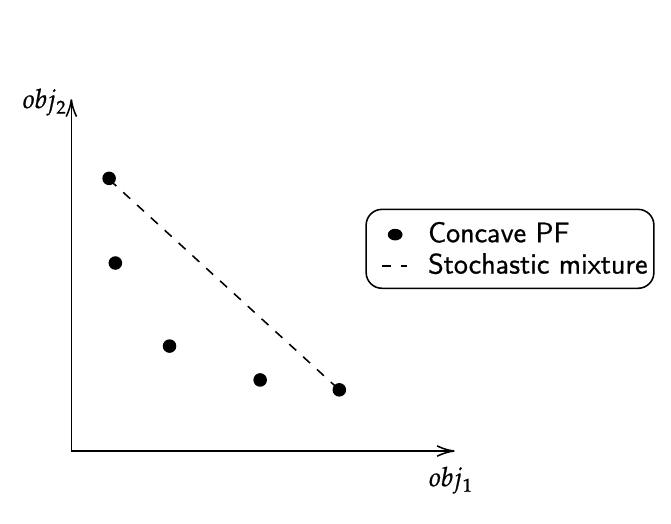}
    \caption{On average, policies resulting from a stochastic mix of deterministic policies dominate the policies in the concave part of the Pareto front.}
    \label{fig:stochastic_mix}
\end{figure}
\noindent When the learned policies are stochastic, the concave part of the PF is dominated by a stochastic mixture of policies~\shortcite{vamplew_constructing_2009}. This can be achieved, for instance, by employing one policy for one episode and another for the subsequent episode (see Figure~\ref{fig:stochastic_mix}). Hence, in such cases, one can restrict to the usage of linear scalarization.

\begin{figure}
    \centering
    \includegraphics[width=0.4\textwidth]{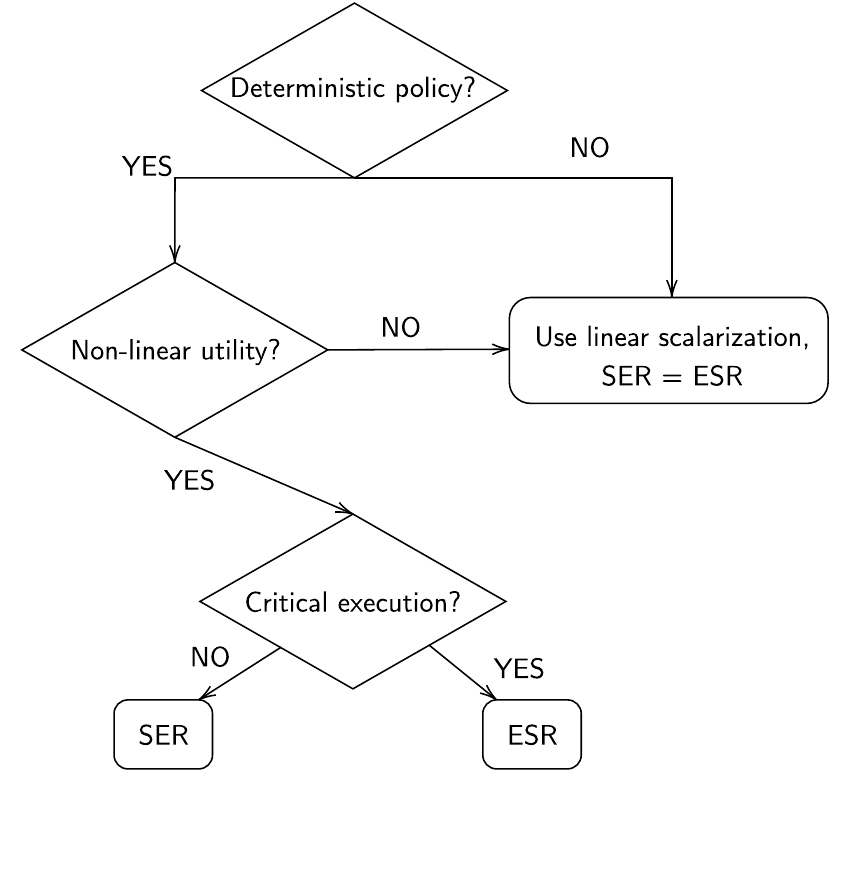}
    \caption{ESR vs. SER decision process.}
    \label{fig:esr_ser}
\end{figure}

Finally, the choice between ESR and SER depends on the criticality of the resulting policy decisions. ESR, where scalarization is applied before the expectation, is suitable for critical applications like cancer detection, as every episodic return is crucial. In contrast, SER applies scalarization on the average return, making it suitable for repetitive applications, such as investments.
 Given these observations, capturing policies in the concave parts of the PF is valuable when the user has a non-linear utility and aims to learn deterministic policies under the ESR criterion.  Figure~\ref{fig:esr_ser} summarizes the decision process to choose between ESR and SER settings in MORL. It is worth noting that in the literature most of the published algorithms optimize for the SER setting or under linear scalarization, leaving ESR understudied~\shortcite{hayes_practical_2022,ropke_distributional_2023}.

\subsubsection{Cooperation}
To improve the sample efficiency compared to the single solution approach (vanilla outer loop), some works rely on cooperation mechanisms where the information gathered by a policy is shared with other policies.  It is worth noting that RL policies are often over-parameterized, and thus, policies having very different parameters may lead to close evaluation points in the objective space. However, in MORL, these parameters are often kept similar. Usually, this is enforced by the fact that policies are initialized to be very close to each other (\textit{e.g.} via transfer learning), or their parameters are kept close to each other via the cooperation mechanisms.

While the neighborhood definitions and exchange triggers from MOO/D can be readily transferred to MORL/D, the knowledge exchange mechanism in MORL/D requires specific attention. This is because policies in MORL/D are typically encoded in regression structures or tables, which is different from the decision variable encoding in MOO/D. This distinction between MOO/D and MORL/D significantly impacts the design of information exchange techniques in the context of multi-objective reinforcement learning. Based on the surveyed articles in the MORL literature, we identify three ways to exchange information between neighbors.

\begin{figure}
    \centering
    \includegraphics[width=0.8\textwidth]{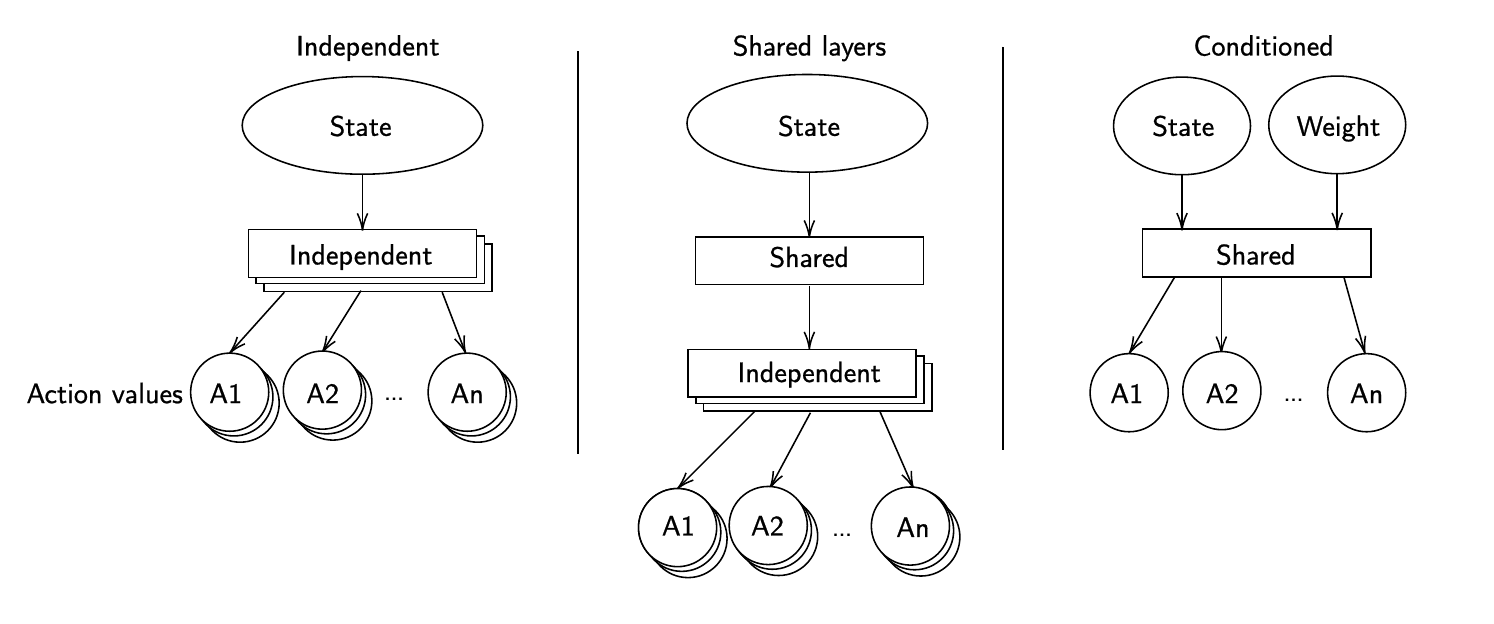}
    \caption{Shared regression structure schemes. Independent policies and conditioned regression lie on both ends of the shared regression spectrum. Independent policies do not share any parameter, whereas conditioned regression allows encoding multiple policies into a single regression structure. Shared layers lie in the middle, sharing part of the network parameters while leaving some to be independent.}
    \label{fig:shared_repr}
\end{figure}

\paragraph{Shared regression structure.} While vanilla outer loop typically relies on completely independent policies (Figure~\ref{fig:shared_repr}, left), some works propose to share information between distinct policies by directly sharing part (or all) of the regression structure. This way, part of the parameters are shared, allowing to learn fewer parameters but also to share gathered experiences with multiple policies at the same time. In this fashion, \shortciteA{chen_combining_2020} proposes to share some base layers between multiple DNNs representing the policies (Figure~\ref{fig:shared_repr}, middle). More extreme cases, such as \shortciteA{abels_dynamic_2019}, propose using only one regression structure by conditioning the input on the weights (Figure~\ref{fig:shared_repr}, right). This approach, often referred to as Conditioned Network, also works for regression trees \shortcite{castelletti_multiobjective_2013}, hence we refer to it as conditioned regression (CR).\footnote{This is similar to what is termed \textit{parameter sharing} in multi-agent RL, \textit{e.g.}~\shortciteA{yu_surprising_2022}.} While it may provide faster convergence to new points in the PF, this technique may forget previously trained policies unless tailored mechanisms are deployed~\cite{abels_dynamic_2019}.


\paragraph{Transfer.} Transfer learning is a common machine learning technique that involves utilizing parameters from a previously trained regression structure to initiate the training of a new regression. In the context of MORL, transfer learning has been applied to various algorithms, allowing the training of a new policy to start from the parameters of the closest trained neighbor policy. This approach has demonstrated improved performance in some MORL algorithms \cite{roijers_point-based_2015,natarajan_dynamic_2005}. 

\paragraph{Shared model.} Another way to exchange information between policies is to share a model of the environment. The idea is that the environment dynamics, \textit{i.e.} $t$ and $\vec r$, are components that need to be estimated, but are the same for each policy that needs to be learned. 

In this fashion, the simplest way to share a model is to use experiences sampled from the environment by one policy to apply Bellman updates to neighbor policies. Such a technique can drastically improve the sample efficiency of MORL methods, yet it is restricted to use an underlying off-policy learning algorithm. This idea is similar to sharing the search memory in MOO/D and is usually achieved by sharing the experience buffer between multiple policies in MORL \shortcite{abels_dynamic_2019,chen_combining_2020}. 

Another technique proposes to use model-based reinforcement learning to learn the dynamics and rewards of the environment in a surrogate model, and to use the learned model to generate samples to learn policies \shortcite{wiering_model-based_2014,alegre_sample-efficient_2023}. Such an approach also improves sample efficiency in MORL algorithms. This setting is very similar to what is called ``same dynamics with different rewards" in model-based RL \cite{moerland_model-based_2023}. 

\subsection{Common Design Choices with RL}

Naturally, existing techniques from RL can also be adapted and applied directly in MORL/D. However, due to the unique multi-objective setting, certain components and aspects of these techniques may need to be modified to accommodate the specific requirements of MORL. This section delves into the design choices and adaptations that originate from the field of RL but are relevant when applied in the context of MORL/D.

\subsubsection{Regression Structure}
As in classical RL, the choice of whether to use a function approximation or a tabular representation depends on the problem the user wants to tackle. Different regression structures allow for different sharing mechanisms. For example, transfer could be applied to any regression structure, but it may be less clear how to effectively share layers in a tabular setting. Independently, such a structure could also be adapted to incorporate multi-objective aspects.

\paragraph{Multi-objective regression structure.} Several authors propose to slightly modify the way information is learned and encoded. Conditioned regression, as discussed above, includes the weight vectors as input to the regression structure. Moreover, it is also possible to vectorize the value function estimator. For example, multi-objective DQN \shortcite{mossalam_multi-objective_2016} outputs $|\mathbb A| \times m$ components, meaning that for each action, it predicts $m$ values corresponding to each objective. Some results suggest that these vectorized value function structures may be more efficient than the scalarized ones~\cite{abels_dynamic_2019}, possibly because the regression structure does not need to capture the scalarization being used when maintaining vectorized estimators.

\subsubsection{Policy Evaluation and Improvement}
In RL, a crucial aspect is the policy improvement step, which typically relies on Bellman update equations. However, standard Bellman updates are designed to handle scalar rewards, whereas MORL deals with vectorized rewards. To adapt the Bellman update to multi-objective settings, various techniques and approaches have been proposed in the MORL literature. 

\paragraph{Scalarized update.}  The simplest approach, called scalarized update, is to use the scalarization function and rely on the Bellman updates from single-objective RL \cite{mossalam_multi-objective_2016,chen_combining_2020}. The main advantage of this approach is that one can rely on the vast RL literature. For example, from an experience tuple $(s_t,a_t,\vec r_t, s_{t+1})$ and a given weight vector $\vec\lambda$, scalarized Q-Learning~\cite{van_moffaert_scalarized_2013} adapts Equation~\ref{eq:q-learn} to the following Bellman update:

\begin{equation*}
\Tilde{q}(s_t,a_t) \leftarrow \Tilde{q}(s_t,a_t) + \alpha\left(\textcolor{myblue}{g(\vec r_t, \vec\lambda)} + \gamma \max_{a'\in \mathbb A} \Tilde{q}(s_{t+1},a') - \Tilde{q}(s_t,a_t)\right).
\end{equation*}

\paragraph{Multi-objective Bellman update.} Other approaches propose to modify the regression function to include more multi-objective aspects. In some cases, such as when relying on a conditioned regression, it makes more sense to optimize the predictions towards a good representation of the full PF. In this sense, \citeA{yang_generalized_2019} proposes a Bellman update called \textit{Envelope}, which optimizes not only across actions for each state but also for multiple weights over a space of preferences $\Lambda$. It relies on the weighted sum scalarization $g^{\text{ws}}$ and stores the q-values as vectors. The update is defined as follows: 

\begin{equation*}
\vec q(s_t,a_t, \vec\lambda) \leftarrow \vec q(s_t,a_t, \vec\lambda) + \alpha\bigg(\vec r_t + \gamma \: \bigg ( \argg_{\vec q}\max_{a'\in \mathbb A, \vec\lambda' \in \Lambda} g^{\text{ws}}\big(\vec q(s_{t+1},a', \vec\lambda'), \vec\lambda\big) \bigg) - \vec{q}(s_t,a_t, \vec\lambda)\bigg),
\end{equation*}

\noindent where the $\argg_{\vec q}\max_{a'\in \mathbb A, \vec\lambda' \in \Lambda}$ operator returns the $\vec q$ vector maximizing over all actions and weights.

\subsubsection{Buffer Strategy}
While experience buffers are common in classical RL to enhance learning efficiency, their utilization in MORL differs somewhat. In MORL, the agent aims to learn multiple policies simultaneously, which can impact how experiences are stored and shared. Hence, the choice of how to manage experience buffers in MORL is influenced by the need to support the learning of multiple policies. 

\paragraph{Replacement.} The classic way to organize an experience buffer is to store experiences based on their recency, where old experiences are discarded for new ones, also called first-in-first-out (FIFO). However, the problem with this kind of reasoning in MORL is that the sequence of experiences sampled by policies on different ends of the PF will probably be very different. Hence, the stored information may not be very useful to improve some policies. 
Thus, this replacement criterion of the buffer could be changed to include multi-objective criteria. Instead of replacing all the experiences in the buffer at each iteration, some elected experiences could stay in the buffer for various iterations. This allows for keeping a diverse set of experiences in the buffer. For example, \citeA{abels_dynamic_2019} proposes to store the return of a sequence of experiences as a signature for the sequence in the replay buffer and a crowding distance operator \cite{deb_fast_2002} is used to keep the diversity of experiences in the buffer.

\paragraph{Selection.} Independently, the way to select which experience to use from the buffer to improve a policy can also include multi-objective criteria. The simplest sampling strategy is to pick experiences from the buffer uniformly. However, this strategy might select a lot of experiences leading to poor learning. As in classical RL, one way to improve the quality of these samples is to rely on priorities \shortcite{schaul_prioritized_2016}. Notably, this technique has been adapted to MORL settings, where each policy keeps its own priority based on its weights \cite{abels_dynamic_2019,alegre_sample-efficient_2023}. 

\paragraph{Neighborhood size.} Instead of having either one buffer for all policies or one buffer per policy, an intermediate granularity of buffers could be shared between neighboring policies. In this way, the experiences in the buffer should be collected by policies that are close to the one being updated. The only existing work that has been found to address this idea is an ablation study presented in \citeA{chen_combining_2020}. 

\subsubsection{Sampling Strategy}

As previously mentioned, the goal of an MORL agent is to learn multiple policies and is generally applied in an offline setting. Thus, in the training stage, there is no such thing as best action for each state, since these can be Pareto incomparable. In addition, various policies will probably lead to the exploration of different areas in the environment. Thus, adapting RL sampling strategies might be beneficial in such cases.

\paragraph{Policy following.} The straightforward adaptation of the RL sampling strategy to MORL is the one implemented in Algorithm~\ref{algo:MORL/D}. At each iteration, the agent chooses one policy to follow and samples the environment according to this policy. 
The choice of which policy to execute at every iteration is very similar to the individual selection problem in MOO/D. Hence, methods coming from MOO/D can be readily applied.  Additionally, it is also possible to use multiple policies in parallel to fill the buffer at each iteration \cite{chen_combining_2020}.

\paragraph{Model-based.} Another approach consists in constructing a model of the environment to systematically sample different areas \cite{moerland_model-based_2023}. This allows for a more global sampling strategy. In this fashion, \shortciteA{felten_metaheuristics-based_2022} proposes to rely on metaheuristics to control the exploration of the agent and to use multi-objective metrics for the exploitation. 

\subsection{Summary}

This section introduced the MORL/D taxonomy and framework, which is based on the integration of RL and MOO/D concepts. The taxonomy allows for the classification and description of existing MORL works while also serving as a guide for new research directions. The framework, with its adaptable nature, enables the direct transfer of knowledge from RL and MOO/D to MORL/D and provides a structured approach for assessing the effectiveness of new ideas.

The following sections explore the practical use of this contribution. First, we classify existing works according to the taxonomy, demonstrating its ability to comprehensively capture the core components of various MORL works. Then, we discuss how to instantiate the framework to address real-world problems. Finally, we provide evidence of the practical application of the framework by conducting experiments on established benchmark problems.

\section{Using MORL/D}

\label{sec:using_morld}

After presenting our taxonomy and framework in the last section, this section illustrates some of its use cases. First, we show how the taxonomy can be used to comprehend existing and future works. Then, we discuss how to make choices about the practical instantiation of the framework to solve a given problem.

\subsection{Classification of Existing Work into our Taxonomy}

This section categorizes three existing works within our taxonomy, demonstrating how the introduced terminology may be applied to position contributions in the broader context of MORL. These works have been chosen because they are representative of typical MORL/D instantiations. For a more exhaustive list of classified works, refer to the table in Appendix~\ref{sec:appendix_table}. Note that reference points are omitted from the classification as all the methods studied in this context rely on linear scalarization, reflecting the current state-of-the-art in MORL.

\subsubsection{PGMORL \shortcite{xu_prediction-guided_2020}}

Prediction-guided MORL (PGMORL)~\cite{xu_prediction-guided_2020} is an evolutionary algorithm that maintains a population of policies learned using a scalarized version of proximal policy optimization (PPO)~\cite{schulman_proximal_2017}. Below, we categorize its components within the MORL/D taxonomy.

\paragraph{Scalarization.} PGMORL relies on the weighted sum scalarization.
\vspace{-2mm}
\paragraph{Weight vectors.} The algorithm generates weight vectors uniformly in the objective space at the beginning of the process.
\vspace{-2mm}
\paragraph{Cooperation.} This algorithm does not enforce cooperation between policies.
\vspace{-2mm}
\paragraph{Selection.}  PGMORL introduces a prediction model that aims to forecast improvements on the PF from previous policy evaluations and specific weight vectors. It uses this predictive model to choose pairs of policies and weight vectors that are expected to result in the most significant predicted improvements, following a tournament-style selection approach. 
\vspace{-2mm}
\paragraph{Archive.} A Pareto archive is employed to maintain snapshots of the best policies over the course of the learning process. Pareto dominance is used as pruning criterion.
\vspace{-2mm}
\paragraph{Regression structure.}  PGMORL maintains a population of both actors and critics, where the critics are multi-objective.
\vspace{-2mm}
\paragraph{Policy improvement.} The algorithm adopts a scalarized variant of PPO~\cite{schulman_proximal_2017} for policy improvement.
\vspace{-2mm}
\paragraph{Buffer.} The buffers in PGMORL operate independently and adhere to the classic PPO strategy, storing experiences based on recency and employing uniform selection.
\vspace{-2mm}
\paragraph{Sampling.} Policy following is the approach adopted in this algorithm. The experiences gathered are used exclusively to train the current policy. This restriction is attributed to the nature of PPO~\cite{schulman_proximal_2017}, which is an on-policy algorithm and offers limited flexibility in terms of sampling strategies.

\subsubsection{Multi-policy SAC \shortcite{chen_combining_2020}}

Multi-policy soft actor-critic~\cite{chen_combining_2020} is a two-stage algorithm that first applies an MORL/D method to learn a set of policies, then uses an evolutionary strategy to finish the search based on the policies found in the first phase. The MORL/D phase keeps a set of policies attached to predefined weight vectors for the entire process. This means that if the user specifies 5 weights, then this phase will return 5 policies. The paragraphs below describe the first phase of the algorithm.

\paragraph{Scalarization.}Multi-policy SAC utilizes the weighted sum scalarization approach. 
\vspace{-2mm}
\paragraph{Weight vectors.} The weight vectors are set manually before the training process starts. These are never changed during the MORL process.
\vspace{-2mm}
\paragraph{Cooperation.} This algorithm incorporates unique cooperation strategies. Firstly, the policies share base layers (Figure~\ref{fig:shared_repr}, middle), facilitating the exchange of information across all policies during the application of gradients in the policy improvement process. Additionally, various experience buffer sharing strategies are explored in the paper, including scenarios where each policy has its own buffer, where each policy shares with adjacent neighbors (those with the closest weight vectors), and where each policy has access to all buffers (a neighborhood that includes all policies).
\vspace{-2mm}
\paragraph{Selection.} At each iteration, all policies are run in a roulette wheel fashion. The weights are kept attached to their policies.
\vspace{-2mm}
\paragraph{Archive.} No archive is used in this algorithm.
\vspace{-2mm}
\paragraph{Regression structure.} This algorithm keeps track of multiple SAC policies~\cite{haarnoja_soft_2018}. 
\vspace{-2mm}
\paragraph{Policy improvement.} The rewards are scalarized before any interaction with the learning process. Hence, the SAC policies are pure single-objective RL.
\vspace{-2mm}
\paragraph{Buffer.} Buffer replacement and selection strategies adhere to conventional practices, recency and uniform selection, respectively. However, this algorithm implements a shared buffer, as explained in the cooperation paragraph above.
\vspace{-2mm}
\paragraph{Sampling.} All policies are used at each iteration for sampling, reflecting a policy following strategy.

\subsubsection{Dynamic weight-conditioned RL~\cite{abels_dynamic_2019}}

Dynamic weight-conditioned RL is a conditioned regression algorithm that models multiple policies into a single regression structure by adding a weight vector as input to the model. This allows such an algorithm to change weight vectors dynamically and thus be employed online but also allows for efficient parameter sharing.

\paragraph{Scalarization.} Weighted sum is used in this algorithm.
\vspace{-2mm}
\paragraph{Weight vectors.} The weight vectors are generated periodically in a random manner.
\vspace{-2mm}
\paragraph{Cooperation.} This algorithm promotes cooperation among multiple policies by sharing parameters using conditioned regression (as illustrated in Figure~\ref{fig:shared_repr}).
\vspace{-2mm}
\paragraph{Selection.} Since all policies are modeled within a single structure, only the weights need to be selected. As explained above, these are randomly generated.
\vspace{-2mm}
\paragraph{Archive.} No Pareto archive is used in this work.
\vspace{-2mm}
\paragraph{Regression structure.} The algorithm keeps one DQN-based conditioned network that outputs multi-objective Q-values for each action.
\vspace{-2mm}
\paragraph{Policy improvement.} The algorithm scalarizes the rewards in each experience according to both the current and historical weights. This approach is employed to facilitate learning for the current weight while preventing the loss of knowledge related to previously learned policies.
\vspace{-2mm}
\paragraph{Buffer.} Various buffer strategies are proposed in this work. First, regarding replacement, it suggests maintaining a diverse set of experiences within a replay buffer, which accelerates learning for a wide range of weight vectors. The diversity criterion is enforced by tracking the return associated with the experience's sequence of actions. A crowding distance operation, equivalent to NSGA-II's~\cite{deb_fast_2002}, is then applied to these returns to quantify buffer diversity. Regarding selection, the authors propose the use of prioritized experience replay, where the priority is computed based on the temporal difference error for the active weight vector and a historical weight vector. For each experience selected for learning, the rewards are scalarized using both the current weight vector and a historical weight vector, akin to the concept of hindsight experience replay in RL~\shortcite{andrychowicz_hindsight_2017}. This is a crucial step to prevent forgetting past policies.
\vspace{-2mm}
\paragraph{Sampling.} Sampling is done by following the policy associated with the current weight.

\subsection{How to Instantiate MORL/D?}
\label{sec:how_to}
As MORL/D presents a combinatorial number of potential instantiations, it may be hard to choose which technique to apply to solve a given problem. This part discusses the decisions of choosing which MORL/D variant to apply given a problem.

First, we provide a formal decision process regarding linear and non-linear scalarization choice in Section~\ref{sec:scalarization_morld} and a graphical representation is illustrated in Figure~\ref{fig:esr_ser}. Additionally, some combinations of techniques do not make sense. For example, conditioned regression may not be combined with transfer learning, and shared layers are not compatible with tabular algorithms. Hence, using common sense on the presented techniques, algorithm designers may be able to restrict the number of possibilities. 

However, for some parts, it is not possible to give absolute directions since a technique that performs well for a given environment might perform poorly for another, as the ``no free lunch" theorem states~\cite{wolpert_no_1997}. Yet, our framework may open ways to solve such an issue by allowing to automate the design of MORL/D algorithms. In such a context, an optimization algorithm is applied to search the space of possible instantiations of MORL/D and decide how to instantiate the parts of the algorithm in order to maximize its performance for a given problem. In RL, these kinds of approaches are referred to as ``AutoRL"~\shortcite{parker-holder_automated_2022,eimer_hyperparameters_2023,felten_hyperparameter_2023}. Hence, we believe our framework could be useful to extend such work to form an AutoMORL solver. Nevertheless, the next section illustrates in practice how we tackle different benchmark problems without such an automated solver at our disposal.

\section{Experiments}
\label{sec:expe}

The last sections introduced the MORL/D framework and taxonomy, and existing works have been classified using the latter. This section shows that the MORL/D framework is not limited to theoretical analysis but is also fit for application. MORL/D has been instantiated to tackle two contrasting multi-objective benchmarks~\shortcite{alegre_mo-gym_2022}, showing how versatile the framework can be. Indeed, the two studied environments contain concave and convex PF, and involve continuous and discrete observations and actions. Each set of experiments has been run 10 times over various seeds for statistical robustness. Additionally, to give a reference, we compare MORL/D results against state-of-the-art methods. In practice, we reuse the data hosted by OpenRLBenchmark~\shortcite{huang_open_2024} to plot metrics from such state-of-the-art methods. MORL/D has been implemented using utilities from the MORL-Baselines~\cite{felten_toolkit_2023} and Pymoo projects~\cite{blank_pymoo_2020}. Our experiments have been run on the high-performance computer of the University of Luxembourg~\shortcite{varrette_management_2014}.\footnote{The code used for experiments is available in MORL-Baselines~\url{https://github.com/LucasAlegre/morl-baselines}.}

\subsection{Assessing Performance}

MORL differs from RL in the way performance results are assessed. Indeed, since multi-policy MORL aims at returning a set of policies and their linked PF, different metrics than episodic return over training time must be used. In general, these metrics aim to turn PFs into scalars to ease comparison. There are two categories of metrics: utility-based metrics, which rely on specific assumptions about the utility function (such as linearity), and axiomatic metrics, which do not make assumptions, but may yield less informative performance information for users~\shortcite{hayes_practical_2022,felten_toolkit_2023}.

\subsubsection{Axiomatic Approaches}
As discussed in Section~\ref{sec:MOO/D}, MOO/D methods involve finding a good approximation of the PF with a focus on two critical aspects: convergence and diversity. To assess convergence, various performance indicators are employed, with the inverted generational distance (IGD) being one such metric \cite{coello_coello_study_2004}. The IGD quantifies the distance between a reference Pareto front, denoted as $\mathcal{Z}$, and the current PF approximation $\mathcal{F}$. Formally, it is computed as follows:

\begin{equation*}
    \mathrm{IGD}(\mathcal{F}, \mathcal{Z}) = \frac{1}{|\mathcal{Z}|}  \sqrt{\sum_{\vec z \in \mathcal{Z}}\min_{\vec{v^\pi} \in \mathcal{F}} \lVert \vec z - \vec{v^\pi} \rVert ^2}.
\end{equation*}

\noindent On the other hand, to evaluate diversity, metrics such as sparsity~\cite{xu_prediction-guided_2020} come into play. Sparsity measures the average squared distance between consecutive points in the PF approximation:

\begin{equation*}
    S(\mathcal{F}) = \frac{1}{|\mathcal{F}|-1} \sum_{j=1}^m\sum_{i=1}^{|\mathcal{F}|-1} (\tilde P_j(i) - \tilde P_j(i+1))^2,
\end{equation*}

\noindent where $\tilde P_j$ is the sorted list for the $j^{th}$ objective values in $\mathcal{F}$, and $\tilde P_j(i)$ is the $i^{th}$ value in this sorted list.

\begin{figure}
    \centering
    \includegraphics[width=0.4\textwidth]{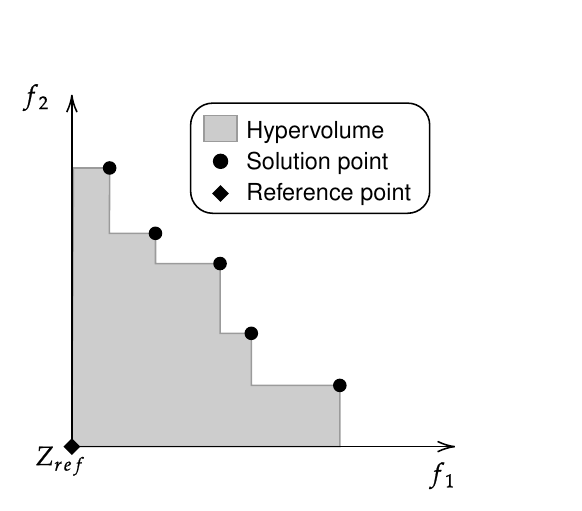}
    \caption{The hypervolume metric in a two objective problem for a given PF.}
    \label{fig:hv}
\end{figure}

Additionally, there are hybrid methods designed to provide insight into both convergence and diversity, such as the hypervolume metric~\cite{zitzler_evolutionary_1999,talbi_metaheuristics_2009}. Hypervolume quantifies the volume of the region formed between each point in the approximated PF and a reference point in the objective space. This reference point, $\vec z_{ref}$, should be carefully chosen as a lower bound for each objective, as illustrated in Figure~\ref{fig:hv}.

Because the Hypervolume metric offers a combined evaluation of both criteria, it is often plotted against the number of timesteps to offer insights into the learning curve within the MORL literature. However, we argue that relying solely on one metric may be insufficient to determine which algorithm outperforms others on a given problem, as illustrated by the experimental results presented later.

\subsubsection{Utility-Based Approaches}

Alternatively, MORL algorithms often seek to maximize the utility of the end user. While the above-mentioned metrics give insights into which methods perform better, they give little information to the end user. To solve such an issue, performance metrics making use of the utility function of the user have also emerged in the fields of MOO and MORL~\cite{talbi_metaheuristics_2009,hayes_practical_2022}. For example, \shortciteA{zintgraf_quality_2015} proposes the expected utility metric (EUM). This unary metric, close to the family of R-metrics in MOO~\cite{talbi_metaheuristics_2009}, is defined as follows:

\begin{equation*}
    \mathrm{EUM}(\mathcal{F})=\mathop{\mathbb{E}}_{\vec \lambda \in \Lambda} \bigg[ \max_{\vec{v^\pi} \in \mathcal{F}} u(\vec{v^\pi}, \vec \lambda)\bigg]
\end{equation*}

\noindent where $u$ represents the utility function of the user (generally weighted sum), that it uses to choose a solution point on the PF, and $\Lambda$ is a set of uniformly distributed weight vectors in the objective space.

\subsection{Benchmark Problems}

\begin{figure}
     \centering
     \begin{subfigure}[b]{0.35\textwidth}
         \centering
         \includegraphics[width=0.85\textwidth]{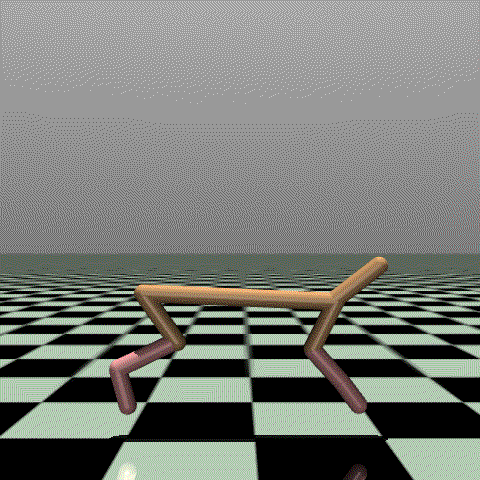}
         \caption{\textit{mo-halfcheetah-v4}.}
         \label{fig:mo-cheetah}
     \end{subfigure}
     \hfill
     \begin{subfigure}[b]{0.35\textwidth}
         \centering
         \includegraphics[width=0.85\textwidth]{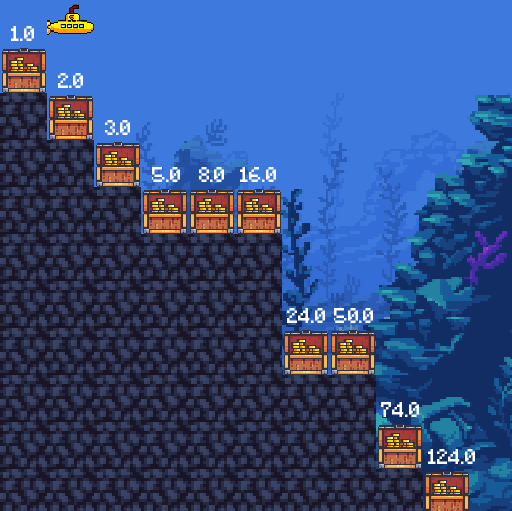}
         \caption{\textit{deep-sea-treasure-concave-v0.}}
         \label{fig:dst_concave}
     \end{subfigure}
     \hfill
    \caption{Studied environments from MO-Gymnasium~\cite{alegre_mo-gym_2022}.}
    \label{fig:envs}
\end{figure}

The first problem, \textit{mo-halfcheetah-v4} (Figure~\ref{fig:mo-cheetah}), presents a multi-objective adaptation of the well-known Mujoco problems~\cite{todorov_mujoco_2012}. In this scenario, the agent takes control of a bipedal robot and aims to mimic the running behavior of a cheetah. The agent's objectives in this environment involve simultaneously maximizing its speed and minimizing its energy consumption. Unlike the original Mujoco implementation, which relies on a weighted sum with hard-coded weights to transform the problem into a single-objective MDP, the multi-objective version treats both objectives independently. Thus, it allows learning various trade-offs for each of these two objectives. 

The second problem of interest is referred to as \textit{deep-sea-treasure-concave-v0} (Figure~\ref{fig:dst_concave}), wherein the agent assumes control of a submarine navigating through a grid-world environment~\shortcite{vamplew_empirical_2011}. In this task, the agent's goal is to collect one of the treasures located on the seabed while striving to minimize the time spent traveling. The rewards for collecting treasures are directly proportional to their distance from the starting point, resulting in conflicting objectives. This benchmark problem holds particular significance due to the presence of a known PF that exhibits a concave shape. This concavity poses a challenge for MORL methods that rely on linear scalarization when the user's objective is to derive a deterministic policy, as discussed in Section~\ref{sec:scalarization_morld}.

\subsection{Solving \textit{mo-halfcheetah-v4}}

This section explains how we tackled the \textit{mo-halfcheetah-v4} problem using MORL/D and compares results against state-of-the-art methods implemented in MORL-Baselines~\cite{felten_toolkit_2023}. 

\subsubsection{MORL/D Variants}
To tackle this task, we initially perform an ablation study for some components of MORL/D (Figure~\ref{fig:morld}). Subsequently, we compare the best version found against various existing MORL algorithms.

The first MORL/D instantiation, which we refer to as MORL/D vanilla, neither performs any cooperation nor weight vector adaptation. We then tried to add cooperation and weight vector adaptation to this vanilla algorithm and examine the results. In practice, we tried adding PSA's method to adapt weights (Equation~\ref{eq:PSA}), and a shared buffer as cooperation mechanism. When relying on one technique, we suffix the technique acronym to MORL/D, \textit{e.g.} MORL/D SB PSA refers to a variant of the algorithm that implements a shared buffer and PSA's weight vector adaptation. It is worth noting that more sophisticated schemes coming from MOO or RL literature can easily be integrated too. Our instantiation is discussed in more detail below.

For this continuous problem, each policy in the population relies on a scalarized multi-objective version of the SAC algorithm~\cite{haarnoja_soft_2018}. Practically, the SAC implementation from \shortciteA{huang_cleanrl_2022} was modified by including multi-objective critics and adding a scalarization function.

\paragraph{Scalarization.} Because the learned policies need not be deterministic, its simplicity to implement, and to avoid making a choice between ESR and SER setting, the weighted sum has been chosen to instantiate MORL/D on this environment.

\vspace{-2mm}
\paragraph{Weight vectors.} For initialization, weight vectors are uniformly generated on a unit simplex using the Riesz s-Energy method from~\citeA{blank_generating_2021}. Moreover, some MORL/D variants perform PSA's weight vector adaptation (Equation~\ref{eq:PSA}) every 50,000 steps.\footnote{In our experience, it is also beneficial to normalize the reward components to facilitate the finding of weight vectors which lead to a diverse PF when the scale between objective values is different.} 

\vspace{-2mm}
\paragraph{Cooperation.} MORL/D vanilla does not implement any cooperation mechanism, whereas MORL/D SB implements shared buffer across the entire population. In the latter, the neighborhood is all other policies in the population and the exchange happens continuously since the buffer is shared by everyone.

\vspace{-2mm}
\paragraph{Selection.} In all variants, each policy is attached to given weights (which can be adapted) and trained using those. To uniformly train all policies, candidates are chosen in a roulette-wheel fashion to sample the experiences.

\vspace{-2mm}
\paragraph{Archive.} To ensure no performance loss after weight adaptation, a Pareto archive has been implemented using the Pareto dominance criterion as sole pruning function. The archive stores snapshots of the SAC policies leading to Pareto optimal points.

\vspace{-2mm}
\paragraph{Regression structure.} The studied MORL/D variants are actor-critic methods based on neural networks. The critics have been modified to implement a multi-objective regression, \textit{i.e.} each critic outputs $m$ values.

\vspace{-2mm}
\paragraph{Policy improvement.}  The scalarization function is applied on the muti-objective estimates from the critics to transform them into scalars. This allows falling back to the original implementation of the Bellman update with scalarized RL.

\vspace{-2mm}
\paragraph{Buffer.} Both MORL/D variants use experience buffers with a recency criterion for storage and sample uniformly from the buffers. In the case of MORL/D SB variants, a single buffer is shared among all the policies. 

\vspace{-2mm}
\paragraph{Sampling strategy.} In all variations, samples are collected by following the selected candidate policy, \textit{i.e.} policy following. Note that policy updates are also performed on the currently followed policy while following the policy, as in the original implementation of SAC.\\

\noindent A set of hyperparameters in both MORL/D and the underlying SAC implementations has been set to conduct our experiments. Table~\ref{tab:hyperparams_sac} in Appendix~\ref{sec:appendix_hps} lists those.

\subsubsection{Experimental Results}

\begin{figure}
    \centering
    \includegraphics[width=0.99\textwidth]{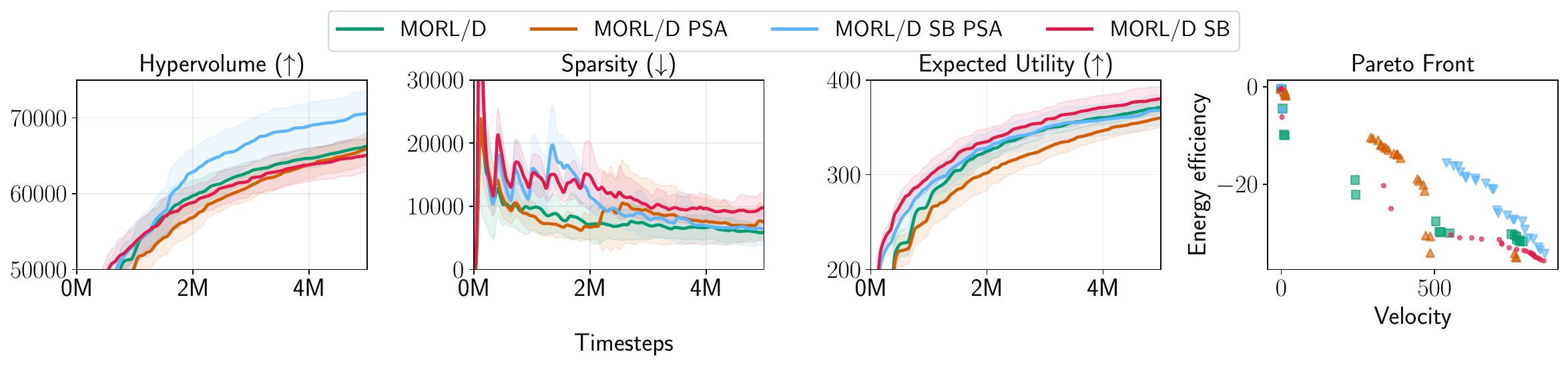}
    \caption{Average and 95\% confidence interval of various metrics over timesteps on \textit{mo-halfcheetah-v4} for variants of MORL/D. The rightmost plot is the resulting PF of each method's best run (best hypervolume).}
    \label{fig:ablation_cheetah}
\end{figure}

\paragraph{Ablation study.} Figure~\ref{fig:ablation_cheetah} shows the results of an ablation study of various versions of MORL/D on the \textit{mo-halfcheetah} problem. A salient observation is that all MORL/D variants consistently improve their Pareto sets and PFs due to the utilization of the Pareto archive. When it comes to hypervolume (reference point $(-100, -100)$), it is evident that MORL/D with a shared buffer and PSA weight adaptation (MORL/D SB PSA) appears to outperform other variants in general. However, in terms of metrics such as sparsity and expected utility, no particular variant emerges as superior to the others.

The right plot of the PF illustrates how policies are distributed across the objective space for their best run.\footnote{It is worth emphasizing that such a plot reflects the performance of only one run, while the metric plots reflect the general performance over multiple runs.} Notably, this graph reveals an unequal scaling of objectives, which implies that scalarized values, based on uniformly spaced weights and metrics like expected utility, may exhibit bias in favor of objectives with larger scales. This phenomenon is corroborated by the fact that most of the Pareto optimal points are located on the right side of the plot, primarily because the velocity objective exhibits a larger scale compared to energy efficiency. Even though we normalize the rewards using a wrapper for this problem, it appears that this normalization is insufficient to learn a continuous PF. Nevertheless, the plot underscores that adding cooperation and weight adaptation to the vanilla MORL/D improves its performance. 

\begin{figure}
    \centering
    \includegraphics[width=0.99\textwidth]{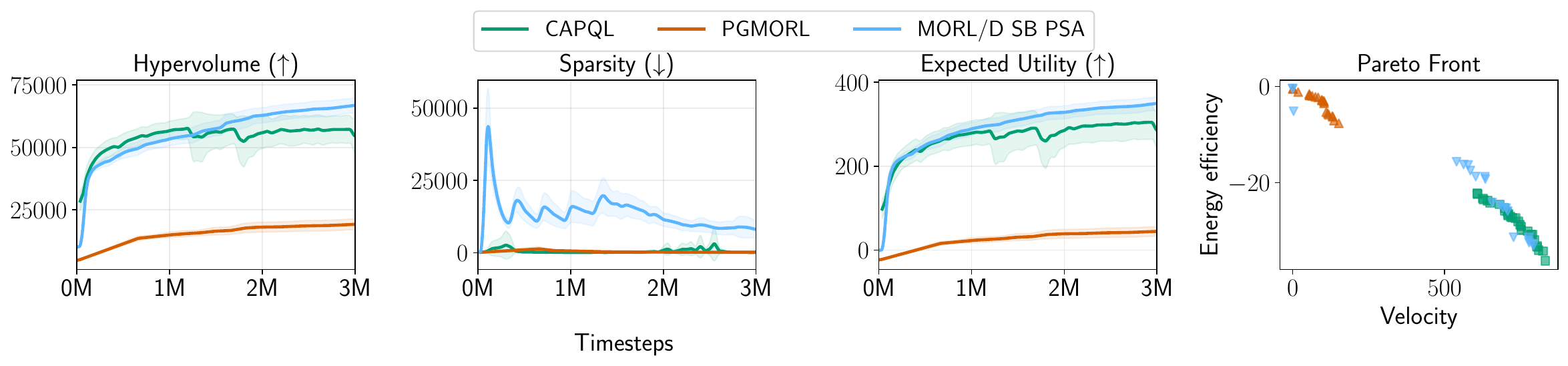}
    \caption{Average and 95\% confidence interval of various metrics over timesteps on \textit{mo-halfcheetah-v4}, MORL/D compared against state-of-the-art methods. The rightmost plot is the PF of each method's best run (best hypervolume).}
    \label{fig:morld_sota}
\end{figure}

\paragraph{MORL/D vs. state of the art.} We now compare MORL/D SB PSA to the state-of-the-art methods in Figure~\ref{fig:morld_sota} to gauge its performance against established baselines. Specifically, we evaluate our algorithm against PGMORL~\cite{xu_prediction-guided_2020} and CAPQL~\cite{lu_multi-objective_2023}. We have previously discussed the first algorithm in Section~\ref{sec:using_morld}. On the other hand, it is interesting to note that CAPQL closely resembles the instantiation of MORL/D we have employed to address this problem. CAPQL, indeed, relies on scalarized SAC using a weighted sum. Nevertheless, it distinguishes itself by relying on conditioned regression and randomly sampling new weight vectors for each environment step.

Based on the data presented in Figure~\ref{fig:morld_sota}, our implementation achieves performance that is comparable to or even superior to the state of the art, particularly in terms of hypervolume and expected utility. This, in particular, contradicts the current belief that conditioned regression-based methods are more efficient than relying on multiple networks~\shortcite{abels_dynamic_2019}. It is worth noting that the performance of CAPQL exhibits occasional drops during its training phase, which we suspect may be attributed to the conditioned neural network employed in the algorithm that forgets previously learned policies. This issue does not arise when using multiple neural networks in conjunction with a Pareto archive. We also provide performance in terms of runtime in Appendix~\ref{appendix:additional_results}.

Additionally, the PF plot reveals a noteworthy observation: while other algorithms tend to produce nearly continuous PFs on one side of the objective space, our algorithm discovers policies on both ends, contributing to improved diversity. However, sparsity appears to be more favorable in the case of the other algorithms. We believe that this highlights a limitation of this state-of-the-art metric: it assesses distance based on the points found by the algorithm rather than considering the entire objective space. Consequently, an algorithm that locates only a few closely clustered points may exhibit a low sparsity score despite providing smaller diversity.

\subsection{Solving \textit{deep-sea-treasure-concave-v0}}

This section illustrates how MORL/D can be used to solve problems involving PFs with concave parts.

\subsubsection{MORL/D Variant}

In this section, our MORL/D algorithm depends on the expected utility policy gradient algorithm (EUPG)~\cite{roijers_multi-objective_2018}. EUPG is a single-policy ESR algorithm able to learn policies with non-linear scalarization. Employing such an algorithm with different weights enables us to capture policies within the concave part of the PF, which is uncommon in the current MORL literature. This section demonstrates how MORL/D can serve as a framework for converting pre-existing single-policy MORL algorithms into multi-policy ones.

\paragraph{Scalarization.} In this case, we are interested in finding the concave points in the PF. Hence, we rely on the Chebyshev function (Section~\ref{sec:MOO/D}), which is a non-linear scalarization.

\vspace{-2mm}
\paragraph{Reference points.} The Chebyshev scalarization necessitates using a utopian reference point $\vec z$. In many cases, this reference point is hard to set in advance. Hence, we propose to automatically adapt it over the course of the learning process by setting $\vec z$ to be the maximum value observed for each objective, plus a factor $\tau = 0.5$. 

\vspace{-2mm}
\paragraph{Weight vectors.} As for \textit{mo-halfcheetah}, the Riesz s-Energy method is used to uniformly generate weight vectors. Moreover, PSA's weight vectors adaptation (Equation~\ref{eq:PSA}) is used every 1,000 steps. 

\vspace{-2mm}
\paragraph{Cooperation.} No cooperation has been implemented on this problem.

\vspace{-2mm}
\paragraph{Selection.} Each policy is attached to given weights (which can be adapted) and trained using those. To uniformly train all policies, candidates are chosen in a roulette-wheel fashion to sample the next experiences.

\vspace{-2mm}
\paragraph{Archive.} A Pareto archive has been implemented using the Pareto dominance criterion as sole pruning function. The archive stores snapshots of the EUPG policies leading to Pareto optimal points.

\vspace{-2mm}
\paragraph{Regression structure.} EUPG relies on a single NN to model the policy. Interestingly, it proposes to condition the NN on the accrued reward to allow learning ESR policies with non-linear scalarization.

\vspace{-2mm}
\paragraph{Policy improvement.}  This MORL/D variant relies directly on EUPG's policy improvement.

\vspace{-2mm}
\paragraph{Buffer.} This algorithm uses experience buffers with a recency criterion for storage and samples uniformly from the buffers. 

\vspace{-2mm}
\paragraph{Sampling strategy.} Similar to the previous algorithm, samples are collected following the selected candidate policy, \textit{i.e.} policy following. Note that policy updates are also performed on the currently followed policy.\\

\noindent The list of hyperparameter values used for these experiments is available in Table~\ref{tab:hyperparams_concave} in Appendix~\ref{sec:appendix_hps}.

\subsubsection{Experimental Results}

\begin{figure}
    \centering
    \includegraphics[width=0.99\textwidth]{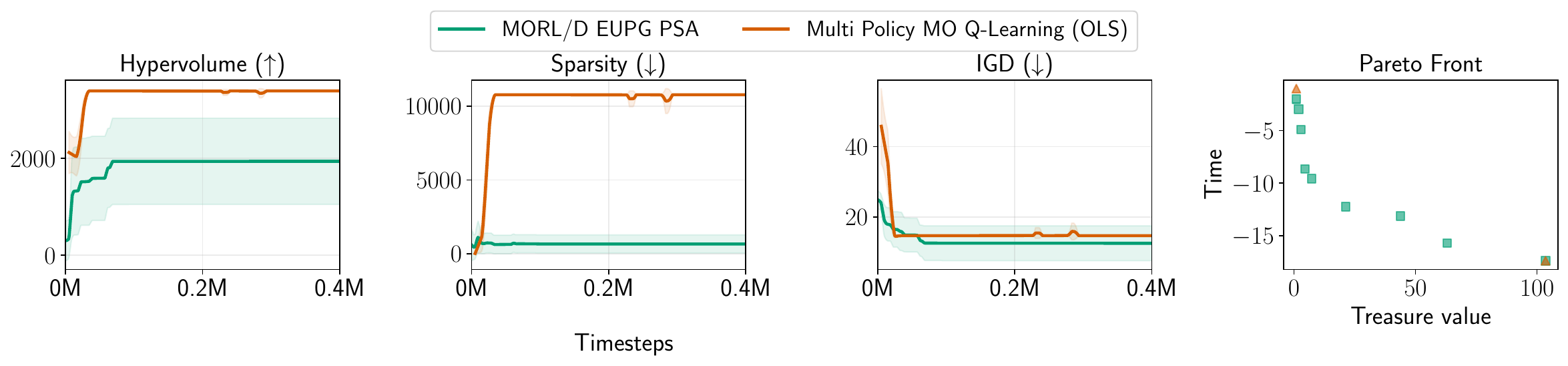}
    \caption{Average and 95\% confidence interval of various metrics over timesteps on \textit{deep-sea-treasure-concave-v0}. The rightmost plot is the resulting PF of each method's best run (best hypervolume).}
    \label{fig:morld_concave}
\end{figure}

Figure~\ref{fig:morld_concave}\ displays the training outcomes of our MORL/D variant, which relies on EUPG as its underlying algorithm. Additionally, we present results for a comparative analysis involving multi-policy multi-objective Q-learning (MPMOQL)~\cite{van_moffaert_scalarized_2013}, a tabular algorithm that depends on multi-objective regression and linear scalarization. Similarly to single-population algorithms in MOO, MPMOQL is run multiple times with various weights generated using optimistic linear support (OLS)~\cite{roijers_point-based_2015} to learn various trade-offs. It is worth emphasizing again that the use of linear scalarization restricts this algorithm from effectively learning points within the concave part of the Pareto front. We used $(0, -50)$ as a reference point for the hypervolume computation. Note that we do not report expected utility in this case, since the metric supposes linear utility, which would not capture any contribution from the concave points in the PF.

Upon examining the plots, it appears that MORL/D performs less favorably in terms of hypervolume compared to MPMOQL. However, when assessing diversity and convergence metrics, a different picture emerges: MORL/D outperforms MPMOQL in terms of sparsity and IGD, respectively. Furthermore, the PF revealed by the best-performing runs clearly demonstrates that MORL/D can capture points within the concave portion of the PF, whereas MPMOQL with linear scalarization can only capture extreme points along the convex hull. 

Upon closer examination, we saw that MPMOQL consistently captures the two extreme points, while MORL/D occasionally struggles to capture points on the right side of the Pareto front, primarily due to limited exploration in EUPG. Notably, the two extreme points captured by MPMOQL result in a very high hypervolume, overshadowing the contributions from the points in the concave region. Consequently, even if MORL/D manages to capture a few points in the concave region, its hypervolume remains lower than that of MPMOQL, which captures only the two extreme points. This underscores the limitation of relying solely on a single performance metric, such as hypervolume or sparsity in the previous example, for algorithm comparison. Indeed, attempting to condense such a wealth of information into a single scalar value comes with inherent trade-offs. Therefore, we advocate for the utilization of multiple metrics and the visualization of PFs whenever possible as a more comprehensive approach.

\section{Future Directions}
This work gave insights on how to transfer existing knowledge from the fields of MOO and RL into MORL. The presented framework allows us to atomically test variations of the algorithm using existing or novel techniques, resulting in novel variations of MORL/D. From our MORL/D taxonomy (Figure~\ref{fig:morld}) and the surveyed articles, we identify key points that have been less studied than others in the next paragraphs.

\paragraph{Non-linear scalarization.} From the surveyed articles and as previously stated in~\citeA{hayes_practical_2022}, a significant portion of MORL algorithms predominantly rely on linear scalarization, thus leaving \textbf{non-linear scalarization} schemes comparatively less explored. This phenomenon might be attributed to the intricate challenges that non-linear scalarization introduces into MORL, as demonstrated by the complexities in the comparison of ESR and SER, see Section~\ref{sec:scalarization_morld}. Moreover, the employment of linear scalarization in many existing works negates the need to \textbf{set or adapt reference points}, leading to this aspect receiving minimal attention in the MORL context. It is also worth noting that, to the best of our knowledge, no existing MORL algorithm has successfully ventured into combining\textbf{ multiple scalarization functions} as has been practiced in the MOO/D domain~\cite{ishibuchi_simultaneous_2010}.

\paragraph{Weight vector adaptation.} In MORL, the generation of weight vectors is often carried out through manual pre-definition or uniform distribution within the objective space, leaving \textbf{weight vector adaptation} schemes being relatively underdeveloped. Notable exceptions can be found in the works of \citeA{roijers_point-based_2015} and \citeA{alegre_sample-efficient_2023}, which employ intelligent weight vector generation based on the current state of the PF. However, once again, it is important to highlight that these methods primarily assume the utilization of linear scalarization. There exist MOO/D methods that are ready to use in MORL/D, as exemplified by PSA (Equation~\ref{eq:PSA}) that we used in this work.


\paragraph{Cooperation schemes.} The MORL/D landscape opens up possibilities for novel cooperation schemes. For instance, when employing multiple regression structures, there exists the potential to infer new policies by combining two existing ones, akin to the utilization of \textbf{crossover operators} in neuroevolution~\shortcite{stanley_designing_2019} or soups of models~\shortcite{wortsman_model_2022}. This approach enables the MORL/D algorithm to blend policy improvements with crossovers, facilitating the efficient generation of a Pareto set of policies.


\paragraph{Parallelizing MORL/D.} The central focus of this article primarily revolves around enhancing sample efficiency, as often seen in RL. Yet, little attention has been given to the enhancement of sample throughput, which holds the potential to substantially reduce the training time required for MORL/D algorithms, even for less sample-efficient (but faster) algorithms.

In this context, maintaining the independence of policies, meaning the enforcement of no cooperation schemes, naturally lends itself to a straightforward approach of breaking down the problem and simultaneously addressing all subproblems in parallel, akin to embarrassingly parallel search ~\cite{regin_embarrassingly_2013}. Though, to the best of our knowledge, the comprehensive exploration of fully parallelized MORL algorithms remains uncharted territory, with no known study having undertaken this multifaceted investigation.

Furthermore, there exists a middle ground where cooperation schemes and parallelized training of multiple policies may offer promising results. However, it is important to note that the sharing of information (cooperation) between subproblems in such scenarios can introduce bottlenecks in parallel programs due to the need for thread synchronization. This represents an exciting avenue for future research and innovation within the field of MORL.

\paragraph{Automated MORL.} Having a modular framework that is instantiable with many techniques from both RL and MOO/D leads to a combinatorial number of choices of instantiation. As discussed in Section~\ref{sec:how_to}, our framework could be combined with automated design techniques to automatically choose well-performing algorithm components for a given problem.


\section{Conclusion}
This work first presented both fields of RL and MOO by breaking up existing solutions into atomic design choices. The differences and common points between these fields of research have been discussed and notable literature is surveyed. 

Subsequently, the paper introduced multi-objective reinforcement learning based on decomposition (MORL/D), a methodology that draws inspiration from both RL and MOO. MORL/D's primary objective is to identify a collection of Pareto optimal policies for solving multi-objective sequential challenges. It employs a scalarization function to break down the multi-objective problem into individual single-objective problems. Building upon the foundations of MOO and RL, the paper introduces a taxonomy focused on solving methods that facilitates the categorization of existing and prospective MORL works. To showcase its utility, a portion of the existing MORL literature has been examined through the taxonomy.

Furthermore, the paper presented a unified framework based on the established taxonomy and adapted it in various ways to conduct experiments on diverse benchmark problems. These experiments demonstrate MORL/D's capacity to address a wide range of challenges through easily identifiable adjustments in the framework's instantiation. Notably, the experiments illustrated how one could port existing knowledge from MOO to MORL, \textit{e.g.} weight vector adaptation while achieving competitive performance compared to current state-of-the-art methods. Moreover, the experimental results and the discussion unveiled important concerns when relying solely on performance metrics to evaluate algorithms.

Lastly, the taxonomy introduced in this work has been utilized to provide insights into potential directions for future MORL research, to leverage knowledge from MOO and RL, or to design entirely novel approaches tailored to the field of MORL.

\acks{
We would like to express our gratitude to the reviewers and editors for their invaluable feedback during the review process. Additionally, we would like to thank Pierre Talbot and Maria Hartmann for proofreading the text.

This work has been funded by the Fonds National de la Recherche Luxembourg (FNR), CORE program under the ADARS Project, ref. C20/IS/14762457.}

\appendix

\section{Reproducibility}
\label{sec:appendix_hps}
This section presents the hyperparameter values used in our experiments. Our code can be found in the MORL-Baselines repository~\cite{felten_toolkit_2023}. Moreover, the results of all runs and hyperparameters are hosted in OpenRLBenchmark~\cite{huang_open_2024}. Environment implementations are available in MO-Gymnasium~\cite{alegre_mo-gym_2022}.

\renewcommand{\arraystretch}{1.3} 
\begin{table}[h]
\centering
\begin{tabular}{l|l|l|}
\cline{2-3}
                                      & \cellcolor[HTML]{D9D9D9}\textbf{Hyperparameter} & \cellcolor[HTML]{D9D9D9}{\color[HTML]{333333} \textbf{Value}}                                                                     \\ \hline
\multicolumn{1}{|l|}{\cellcolor[HTML]{D9D9D9}\textbf{MORL/D}} & Population size                    & $n=6$                                                                                  \\ \hline
                                      & Episodes for policy evaluation     & 5                                                                                 \\ \cline{2-3}  
                                      & Update passes                      & $u=10$                                                                             \\ \cline{2-3}
                                      & Total environments steps &                $5e6$                                          \\ \cline{2-3}
                                       & Exchange trigger                 & Every 50,000 steps                                  \\ \cline{2-3}
                                      & Scalarization                  & Weighted sum                                               \\ \cline{2-3}
                                      & Weight adaptation                  & PSA, $\delta=1.1$                                                                      \\ \hline 
\multicolumn{1}{|l|}{\cellcolor[HTML]{D9D9D9}\textbf{SAC}}    & Buffer size                        & $10^6$                                                                          \\ \hline
                                      & Gamma                              & 0.99                                                                               \\ \cline{2-3} 
                                      & Target smoothing coefficient       & 0.005                                                                              \\ \cline{2-3} 
                                      & Batch size                         & 256                                                                                \\ \cline{2-3} 
                                      & Steps before learning              & 15,000                                                                             \\ \cline{2-3} 
                                      & Hidden neurons in Neural Networks      & {[}256;256{]}  
                                      \\ \cline{2-3} 
                                      & Activation function in Neural Networks      & {ReLU}  
                                      \\ \cline{2-3} 
                                      & Actor learning rate                & $3e{-4}$                                                                             \\ \cline{2-3} 
                                      & Critic learning rate               & $10^{-3}$                                                                              \\ \cline{2-3} 
                                      & Actor training frequency           & 2                                                                                  \\ \cline{2-3} 
                                      & Target training frequency          & 1                                                                                  \\ \cline{2-3} 
                                      & Entropy regularization coefficient & automatic                                                                          \\ \cline{2-3} 
\end{tabular}
\caption{Hyperparameters for MORL/D on \textit{mo-halfcheetah-v4}.}
\label{tab:hyperparams_sac}
\end{table}
\renewcommand{\arraystretch}{1.} 

\renewcommand{\arraystretch}{1.3} 
\begin{table}[]
\centering
\begin{tabular}{l|l|l|}
\cline{2-3}
                                      & \cellcolor[HTML]{D9D9D9}\textbf{Hyperparameter} & \cellcolor[HTML]{D9D9D9}{\color[HTML]{333333} \textbf{Value}}                                                                     \\ \hline
\multicolumn{1}{|l|}{\cellcolor[HTML]{D9D9D9}\textbf{MORL/D}} & Population size                    & $n=10$                                                                                  \\ \hline
                                      & Episodes for policy evaluation     & 5                                                                                 \\ \cline{2-3}  
                                      & Update passes                      & $u=10$                                                                             \\ \cline{2-3}
                                      & Total environments steps  &               $4e5$                                          \\ \cline{2-3}
                                       & Exchange trigger                 & Every 1,000 steps     \\ \cline{2-3}
                                      & Scalarization                  & Chebyshev 
                                      \\ \cline{2-3}
                                      & Optimization criterion                  & ESR  
                                      \\ \cline{2-3}
                                      & Weight adaptation                  & PSA: $\delta=1.1$                                                                      \\ \hline 
\multicolumn{1}{|l|}{\cellcolor[HTML]{D9D9D9}\textbf{EUPG}}    & Buffer size                        & $10^5$                                                                          \\ \hline
                                      & Gamma                              & 0.99                                                                               \\ \cline{2-3} 
                                      & Hidden neurons in NN      & {[}32;32{]}  
                                      \\ \cline{2-3} 
                                      & Activation function in NN      & {Tanh}  
                                      \\ \cline{2-3} 
                                      & Learning rate                & $1e{-4}$                                                            \\ \cline{2-3} 
\end{tabular}
\caption{Hyperparameters for MORL/D on \textit{deep-sea-treasure-concave-v0}.}
\label{tab:hyperparams_concave}
\end{table}
\renewcommand{\arraystretch}{1.} 

\section{Additional Works Classified in our Taxonomy}
\label{sec:appendix_table}
\afterpage{%
    \clearpage

    \begin{landscape}
    {
    \setlength{\tabcolsep}{0.5em} 
    \renewcommand{\arraystretch}{1.2} 
        \begin{table*}
        \centering
        \footnotesize
        \begin{tabular}{p{2cm}|llll|l||l|lll|l}
        \cline{2-11} 
        \multicolumn{1}{l|}{}                                                                                                                                                                                                                                                                                                                                                        & \multicolumn{5}{c||}{\textbf{MOO}}                                                                                                                                                                                                                                 & \multicolumn{5}{c|}{\textbf{RL}}                                                                                                                                                                               \\                                                              \hline
        \multicolumn{1}{|l|}{\cellcolor[HTML]{D9D9D9}}                                                                       & \multicolumn{2}{c|}{\cellcolor[HTML]{D9D9D9}\textbf{Weight vectors}}                                                                                & \multicolumn{3}{c||}{\cellcolor[HTML]{D9D9D9}\textbf{Cooperation}}                                                                                                                                                                                    & \multicolumn{1}{l|}{\cellcolor[HTML]{D9D9D9}}                                                                        & \multicolumn{1}{l|}{\cellcolor[HTML]{D9D9D9}}                                                  & \multicolumn{2}{c|}{\cellcolor[HTML]{D9D9D9}\textbf{Buffer}}                                                                                                                 & \multicolumn{1}{l|}{\cellcolor[HTML]{D9D9D9}}                                    \\ \cline{2-6} \cline{9-10}
\multicolumn{1}{|c|}{\multirow{-2}{*}{\cellcolor[HTML]{D9D9D9}\textbf{Reference}}}                                                                                  & \multicolumn{1}{l|}{\textbf{When?}} & \multicolumn{1}{l|}{\textbf{How?}}                                                                            & \multicolumn{1}{l|}{\textbf{Neighb.}}                   & \multicolumn{1}{l|}{\textbf{Mechanism}}                                                        & \multicolumn{1}{l||}{\textbf{Trigger}}    & 
\multicolumn{1}{c|}{\multirow{-2}{*}{\cellcolor[HTML]{D9D9D9}\begin{tabular}[c]{@{}l@{}}\textbf{Regression} \\ \textbf{structure}\end{tabular}}}                             & 
\multicolumn{1}{c|}{\multirow{-2}{*}{\cellcolor[HTML]{D9D9D9}\begin{tabular}[c]{@{}l@{}}\textbf{Policy} \\ \textbf{improv.}\end{tabular}}}      & 

\multicolumn{1}{l|}{\textbf{Neighb.}}                                                               & \textbf{\begin{tabular}[c]{@{}l@{}}\textbf{Storage \&} \\ \textbf{Sampling} \\ \textbf{Strategy}\end{tabular}}                                                 &
\multicolumn{1}{c|}{\multirow{-2}{*}{\cellcolor[HTML]{D9D9D9}\begin{tabular}[c]{@{}l@{}}\textbf{Sampling} \\ \textbf{strategy}\end{tabular}}} \\ \hline

\multicolumn{1}{|p{2cm}|}{\shortciteA{roijers_computing_2015}}                                      & \multicolumn{1}{l|}{Dynamic}        & \multicolumn{1}{l|}{\begin{tabular}[c]{@{}l@{}}Adaptive - \\ OLS\end{tabular}} & \multicolumn{1}{l|}{\begin{tabular}[c]{@{}l@{}}Single - \\ Closest weight\end{tabular}}                                                    & \multicolumn{1}{l|}{Transfer}                                                                  & Periodic & $n \times $ Tabular                                                               & \multicolumn{1}{l|}{\begin{tabular}[c]{@{}l@{}}Scalarized \\ POMDP \\ solver\end{tabular}}        & \multicolumn{1}{l|}{/}                                                                         & /                                                           & \multicolumn{1}{l|}{\begin{tabular}[c]{@{}l@{}}Policy \\ following\end{tabular}}                                            \\ \hline
\multicolumn{1}{|p{2cm}|}{\shortciteA{mossalam_multi-objective_2016}}                                         & \multicolumn{1}{l|}{Dynamic}        & \multicolumn{1}{l|}{\begin{tabular}[c]{@{}l@{}}Adaptive - \\ OLS \end{tabular}}                                                                                      & \multicolumn{1}{l|}{\begin{tabular}[c]{@{}l@{}}Single - \\ Closest weight\end{tabular}}                                                    & \multicolumn{1}{l|}{Transfer}                                                                  & Periodic & \begin{tabular}[c]{@{}l@{}}$n \times$ DNN\\ + MO reg.\end{tabular} & \multicolumn{1}{l|}{\begin{tabular}[c]{@{}l@{}}Scalarized\\ DQN\end{tabular}}                  & \multicolumn{1}{l|}{Indep.}                                                                         & \begin{tabular}[c]{@{}l@{}}Recency + \\ Uniform\end{tabular}                                                            & \multicolumn{1}{l|}{\begin{tabular}[c]{@{}l@{}}Policy \\ following\end{tabular}}                                            \\ \hline
\multicolumn{1}{|p{2cm}|}{\citeA{chen_combining_2020}}                                         & \multicolumn{1}{l|}{Static}         & \multicolumn{1}{l|}{Manual}                                                                                   & \multicolumn{1}{l|}{ All} & \multicolumn{1}{l|}{\begin{tabular}[c]{@{}l@{}}Shared buffer\\ Shared layers\end{tabular}} & Continuous                 & $n \times$ DNN                                                                     & \multicolumn{1}{l|}{\begin{tabular}[c]{@{}l@{}}Scalarized\\ SAC\end{tabular}}                  & \multicolumn{1}{l|}{All} & \begin{tabular}[c]{@{}l@{}}Recency + \\ Uniform\end{tabular}                                                           & \multicolumn{1}{l|}{\begin{tabular}[c]{@{}l@{}}Parallel \\ policy \\ following\end{tabular}}                                   \\ \hline
\multicolumn{1}{|p{2cm}|}{\shortciteA{yang_generalized_2019} }                                      & \multicolumn{1}{l|}{Dynamic}        & \multicolumn{1}{l|}{Random}                                                                          & \multicolumn{1}{l|}{All}                                                                        & \multicolumn{1}{l|}{CR}                                                                  & Continuous                 & $1 \times$ DNN                                                                    & \multicolumn{1}{l|}{\begin{tabular}[c]{@{}l@{}}Envelope \\ DQN\end{tabular}}                   & \multicolumn{1}{l|}{All}                                                                                 & \begin{tabular}[c]{@{}l@{}}HER + \\ Recency + \\ Uniform\end{tabular}                                                            & \multicolumn{1}{l|}{\begin{tabular}[c]{@{}l@{}}Policy \\ following\end{tabular}}                                            \\ \hline
\multicolumn{1}{|p{2cm}|}{\shortciteA{xu_prediction-guided_2020}}                     & \multicolumn{1}{l|}{Dynamic}        & \multicolumn{1}{l|}{Uniform}                                                                                  & \multicolumn{1}{l|}{None}                                                                       & \multicolumn{1}{l|}{None}                                                                      & None                         & \begin{tabular}[c]{@{}l@{}}$n \times$ DNN\\ + MO reg.\end{tabular}                                                                 & \multicolumn{1}{l|}{\begin{tabular}[c]{@{}l@{}}Scalarized \\ PPO\end{tabular}}                 & \multicolumn{1}{l|}{Indep.}                                                                         & \begin{tabular}[c]{@{}l@{}}Recency + \\ Uniform\end{tabular}                                                           & \multicolumn{1}{l|}{\begin{tabular}[c]{@{}l@{}}Policy \\ following\end{tabular}}                                            \\ \hline
\multicolumn{1}{|p{2cm}|}{\citeA{abels_dynamic_2019} }                                                 & \multicolumn{1}{l|}{Dynamic}        & \multicolumn{1}{l|}{Random}                                                                                   & \multicolumn{1}{l|}{All}                                                                        & \multicolumn{1}{l|}{CR}                                                                  & Continuous                 & \begin{tabular}[c]{@{}l@{}}$1 \times$ DNN\\ + MO reg. \end{tabular} & \multicolumn{1}{l|}{\begin{tabular}[c]{@{}l@{}}Scalarized,\\ Multi-weights\\ DQN\end{tabular}} & \multicolumn{1}{l|}{All}                                                                                 & \begin{tabular}[c]{@{}l@{}}HER + \\ PER\\ (Diversity) \end{tabular} & \multicolumn{1}{l|}{\begin{tabular}[c]{@{}l@{}}Policy \\ following\end{tabular}}                                            \\ \hline
\multicolumn{1}{|p{2cm}|}{\citeA{alegre_sample-efficient_2023}}                                                       & \multicolumn{1}{l|}{Dynamic}        & \multicolumn{1}{l|}{\begin{tabular}[c]{@{}l@{}}Adaptive -\\ GPI-LS\end{tabular}}                                                                                   & \multicolumn{1}{l|}{All}                                                                        & \multicolumn{1}{l|}{\begin{tabular}[c]{@{}l@{}}CR\\ Shared model\end{tabular}}                                                                        & Continuous                 & \begin{tabular}[c]{@{}l@{}}$1 \times$ DNN\\ + MO reg. \end{tabular}                                                       & \multicolumn{1}{l|}{\begin{tabular}[c]{@{}l@{}}Scalarized,\\ Multi-weights\\DQN or TD3\end{tabular}}                  & \multicolumn{1}{l|}{All}                                                                                 & \begin{tabular}[c]{@{}l@{}}HER + \\PER\\ (GPI) \end{tabular}                                                          & \multicolumn{1}{l|}{\begin{tabular}[c]{@{}l@{}}Policy \\ following\end{tabular}}                                                      \\ \hline
\multicolumn{1}{|p{2cm}|}{\citeA{castelletti_multiobjective_2013}}                                                       & \multicolumn{1}{l|}{Dynamic}        & \multicolumn{1}{l|}{Random}                                                                                   & \multicolumn{1}{l|}{All}                                                                        & \multicolumn{1}{l|}{CR}                                                                        & Continuous                 & $1 \times$ Trees                                                        & \multicolumn{1}{l|}{\begin{tabular}[c]{@{}l@{}}Scalarized\\ FQI\end{tabular}}                  & \multicolumn{1}{l|}{/}                                                                                 & /                                                           & \multicolumn{1}{l|}{\begin{tabular}[c]{@{}l@{}}Historical \\ dataset\end{tabular}}                                                      \\ \hline
        \end{tabular}
        \caption{A non-exhaustive list of MORL works classified according to the MORL/D taxonomy. OLS = Optimistic Linear Support, DNN = Deep Neural Network, POMDP = Partially Observable MDP, MO reg. = Multi-objective regression, CR = Conditioned Regression, HER = Hindsight Experience Replay, PER = prioritized experience replay, GPI = Generalized Policy Improvement.}
        \label{tab:MORLD-variants}
        \end{table*}}
\end{landscape}
    \clearpage
}

Table~\ref{tab:MORLD-variants} presents existing MORL work classified according to the proposed MORL/D taxonomy (Figure~\ref{fig:morld}). This demonstrates the modularity of the proposed framework and offers a high-level view of the current state of the art in MORL. The table is separated by a double vertical line showing again the boundaries between traits inherited from MOO and RL. For each trait, a column presents the instantiation choice made in each paper. 

For instance, in the first line, the work of \citeauthor{roijers_multi-objective_2017} dynamically assigns weight vectors based on Optimistic Linear Support (OLS). As a cooperation mechanism, the algorithm reuses (transfers) the knowledge from the closest already trained policy to hot-start the training of a new policy. The algorithm relies on $n$ tabular representations, where $n$ is the number of desired policies. The Bellman update used is based on a Partially Observable MDP solver (POMDP). Finally, the sampling strategy proposes to follow the policy currently being trained (with its internal exploration technique).

All the works referenced in the table make use of the weighted sum scalarization. Thus, scalarization and reference point columns have been omitted from the table since they would bring little information. In general, non-linear scalarization schemes are understudied when compared to the weighted sum. Additionally, for space reasons, population selection and archive have not been represented either. In both traits, the work of \shortciteA{xu_prediction-guided_2020} is particularly interesting, as explained in Section~\ref{sec:using_morld}.

\section{Additional Results}

\label{appendix:additional_results}
\begin{figure}
    \centering
    \includegraphics[width=0.8\textwidth]{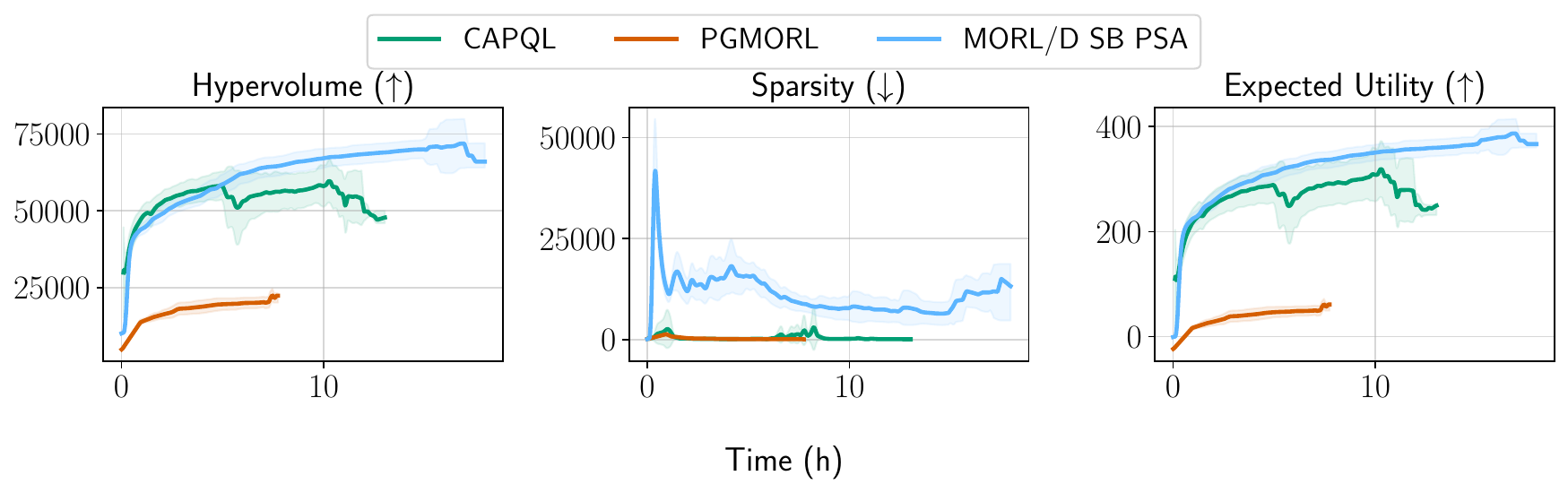}
    \caption{Comparisons in terms of runtime for the \textit{mo-halfcheetah-v4}.}
    \label{fig:morld_cheetah_time}
\end{figure}

\begin{figure}
    \centering
    \includegraphics[width=0.8\textwidth]{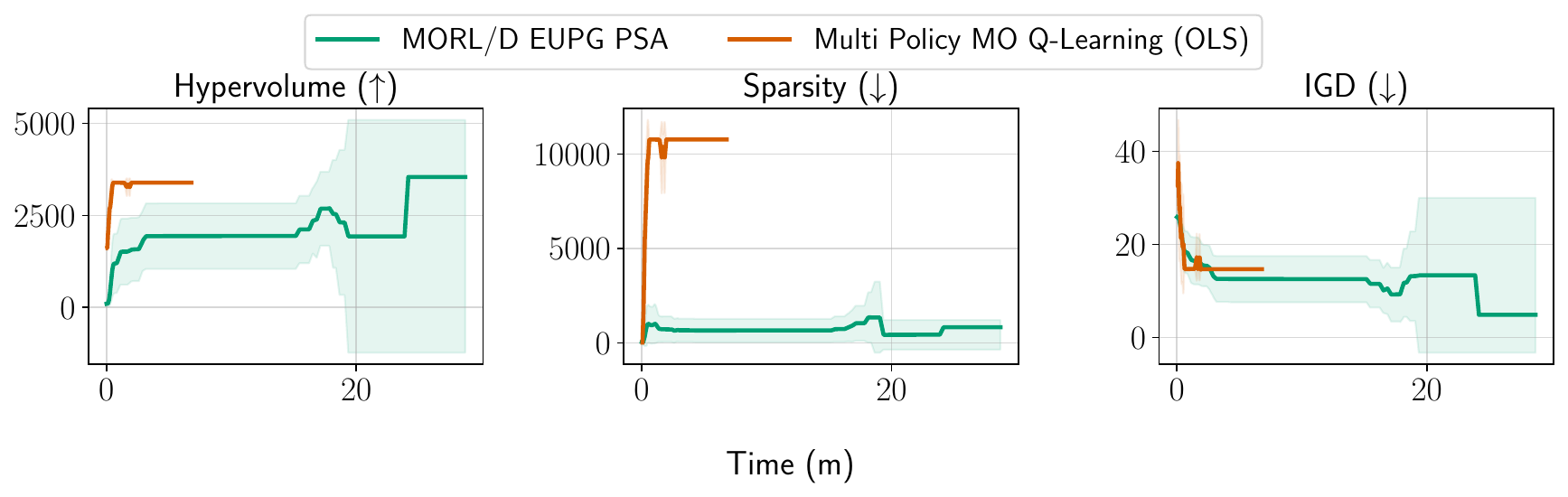}
    \caption{Comparisons in terms of runtime for the \textit{deep-sea-treasure-concave-v0}.}
    \label{fig:morld_concave_time}
\end{figure}

Figures~\ref{fig:morld_cheetah_time} and~\ref{fig:morld_concave_time} give results in terms of runtime of algorithms. Similar conclusions to what has been discussed in the main paper can be drawn here too.

\bibliography{references}
\bibliographystyle{theapa}

\end{document}